\documentclass{article}

\makeatletter

\newlength{\fboxhsep}
\newlength{\fboxvsep}

\newlength{\fboxtoprule}
\newlength{\fboxbottomrule}
\newlength{\fboxleftrule}
\newlength{\fboxrightrule}

\def\@frameb@xother#1{%
  \@tempdima\fboxtoprule
  \advance\@tempdima\fboxvsep
  \advance\@tempdima\dp\@tempboxa
  \hbox{%
    \lower\@tempdima\hbox{%
      \vbox{%
        \hrule\@height\fboxtoprule
        \hbox{%
          \vrule\@width\fboxleftrule
          #1%
          \vbox{%
            \vskip\fboxvsep
            \box\@tempboxa
            \vskip\fboxvsep}%
          #1%
          \vrule\@width\fboxrightrule}%
        \hrule\@height\fboxbottomrule}%
    }%
  }%
}

\long\def\fboxother#1{%
  \leavevmode
  \setbox\@tempboxa\hbox{%
    \color@begingroup
    \kern\fboxhsep{#1}\kern\fboxhsep
    \color@endgroup}%
  \@frameb@xother\relax}

\makeatother

\PassOptionsToPackage{numbers, compress}{natbib}
\usepackage{natbib}
\usepackage{multirow}
\usepackage{listings}
\usepackage{xcolor}
\usepackage{placeins}
\usepackage{amsmath}
\usepackage{color, colortbl}
\usepackage[normalem]{ulem}

\definecolor{codegreen}{rgb}{0,0.6,0}
\definecolor{codeblue}{rgb}{0,0,0.6}
\definecolor{codegray}{rgb}{0.5,0.5,0.5}
\definecolor{codepurple}{rgb}{0.58,0,0.82}
\definecolor{backcolour}{rgb}{0.95,0.95,0.92}
\definecolor{rowcolor}{gray}{0.9}
\definecolor{lightblue}{HTML}{84C7F9}
\definecolor{lighterblue}{HTML}{D4ECFF}

\lstdefinestyle{mystyle}{
    backgroundcolor=\color{backcolour},   
    commentstyle=\color{codegreen},
    keywordstyle=\color{magenta},
    numberstyle=\tiny\color{codegray},
    stringstyle=\color{codepurple},
    basicstyle=\ttfamily\footnotesize,
    breakatwhitespace=false,         
    breaklines=true,                 
    captionpos=b,                    
    keepspaces=true,                 
    numbers=left,                    
    numbersep=5pt,                  
    showspaces=false,                
    showstringspaces=false,
    showtabs=false,                  
    tabsize=2
}

\usepackage[final]{neurips_2022}

\usepackage[utf8]{inputenc} 
\usepackage[T1]{fontenc}    
\usepackage[pagebackref=true]{hyperref} 
\renewcommand*\backref[1]{\ifx#1\relax \else (Cited on page #1) \fi}
\hypersetup{
    colorlinks=true,
    citecolor=blue,
    linkcolor=blue,
    filecolor=magenta,      
    urlcolor=blue,
    pdfborderstyle={/S/U/W 1},
}
\newcommand{\uhref}[2]{\href{#1}{\uline{#2}}}

\usepackage{url}            
\usepackage{booktabs}       
\usepackage{amsfonts}       
\usepackage{nicefrac}       
\usepackage{microtype}      
\usepackage{xcolor}         
\usepackage{graphicx}
\usepackage{wrapfig}

\usepackage{tcolorbox}

\newcommand{\DFRVAL}{DFR$_\text{Tr}^\text{Val}$~}
\newcommand{\DFR}{DFR WGA}
\newcommand{\DFRS}{DFR $s$-WGA}

\newtcolorbox{mybox}{colback=lighterblue,colframe=lightblue}

\title{On Feature Learning in the Presence of Spurious Correlations}

\author{
\normalsize \textbf{
Pavel Izmailov\thanks{Equal contribution.}~\quad
Polina Kirichenko$^*$\quad
Nate Gruver$^*$\quad
Andrew Gordon Wilson} \\
\normalsize New York University
}

\begin{document}

\maketitle
\setcounter{footnote}{0}

\begin{abstract}
\noindent 
Deep classifiers are known to rely on spurious features — patterns which are correlated with the target on the training data but not inherently relevant to the learning problem, such as the image backgrounds when classifying the foregrounds.
In this paper we evaluate the amount of information about the core (non-spurious) features that can be decoded from the representations learned by standard empirical risk minimization (ERM) and specialized group robustness training. 
Following recent work on Deep Feature Reweighting (DFR), we evaluate the feature representations by re-training the last layer of the model on a held-out set where the spurious correlation is broken.
On multiple vision and NLP problems, we show that the features learned by simple ERM are highly competitive with the features learned by specialized group robustness methods targeted at reducing the effect of spurious correlations.
Moreover, we show that the quality of learned feature representations is greatly affected by the design decisions beyond the training method, such as the model architecture and pre-training strategy.
On the other hand, we find that strong regularization is not necessary for learning high quality feature representations.
Finally, using insights from our analysis, we significantly improve upon the best results reported in the literature on the popular Waterbirds, CelebA hair color prediction and WILDS-FMOW problems, achieving 97\%, 92\% and 50\% worst-group accuracies, respectively.
\end{abstract}

\section{Introduction}

In classification problems, a feature is \textit{spurious} if it is predictive of the label without being causally related to it. Models that exploit the predictive power of spurious features can achieve strong average performance on training and in-distribution test data but often perform poorly on sub-groups of the data where the spurious correlation does not hold \citep{geirhos2020shortcut}.
For example, neural networks trained on ImageNet are known to rely on backgrounds \citep{xiao2020noise} or texture \citep{geirhos2018imagenet}, which are often correlated with labels without being causally significant.
Similarly in natural language processing, models often rely on specific words and syntactic heuristics when predicting the sentiment of a sentence or the relationship between a pair of sentences \citep{mccoy2019right, gururangan2018annotation}.
In extreme cases, neural networks completely ignore task-relevant \textit{core} 
features and only use spurious features in their predictions \citep{zech2018variable, shah2020pitfalls}, achieving zero accuracy on the subgroups of the data where the spurious correlation does not hold.

In recent work, \citet{kirichenko2022last} showed that, surprisingly, standard Empirical Risk Minimization~(ERM) learns a high-quality representation of the core features on datasets with spurious correlations, even when the model primarily relies on spurious features to make predictions.
Moreover, they showed that it is often possible to recover state-of-the-art performance on benchmark 
spurious correlation 
problems by simply retraining the last layer of the model on a small held-out dataset where the spurious correlation does not hold. This procedure is called Deep Feature Reweighting (DFR).

In this paper, we provide an in-depth study of the factors that affect the \textit{quality of learned representations} in the presence of spurious correlations:
how accurately we can decode the core features from the learned representations.
Following \citet{kirichenko2022last}, we break the problem of training a robust classifier into two tasks: extracting feature representations and training a linear classifier on these features.
In order to study the feature learning in isolation, we use the DFR procedure to learn an optimal linear classifier on the feature representations, and evaluate the features learned with different training methods, neural network architectures, and hyper-parameters.

First, on a range of problems with spurious correlations we show that while specialized group robustness methods such as group distributionally robust optimization (group DRO)
\citep{sagawa2019distributionally} can significantly outperform the standard ERM training, the quality of the features learned by ERM is highly competitive:
by applying the DFR procedure to the features learned by ERM and group DRO we achieve similar performance.
Furthermore, we show that the performance improvements of group DRO are largely explained by the better weighting of the learned features in the last classification layer, and not by learning a better representation of the core features.
This observation has high practical significance, as the problem of training the last layer of the model is much simpler both conceptually and computationally than training the full model to avoid spurious correlations
\citep{kirichenko2022last}.

Next, focusing on the ERM training, we explore the effect of model class, pretraining strategy and regularization on feature learning.
We find a linear dependence between the in-distribution accuracy of the model and the worst group accuracy after applying DFR, meaning that on natural datasets good generalization typically implies good feature learning, even in the presence of spurious features.
Further, we show that the pre-training strategy has a very significant effect on the quality of the learned features, while strong regularization does not significantly improve the feature representations on most benchmarks.

Finally, by finetuning a pretrained state-of-the-art ConvNext model~\citep{liu2022convnet},
we significantly outperform the best reported results on the popular Waterbirds \citep{sagawa2019distributionally}, CelebA hair color and WILDS FMOW \citep{koh2021wilds} spurious correlation benchmarks, using only simple ERM training followed by DFR.

Our code is available at \href{https://github.com/izmailovpavel/spurious_feature_learning}{\url{github.com/izmailovpavel/spurious_feature_learning}}.

\section{Related Work}

Numerous works describe how neural networks can rely on spurious correlations in real world problems.
In vision, neural networks can learn to rely on an image's background \citep{xiao2020noise, sagawa2019distributionally, moayeri2022comprehensive},
secondary objects \citep{kolesnikov2016improving, rosenfeld2018elephant, singla2021salient, shetty2019not, alcorn2019strike, mo2021object},
object textures \citep{geirhos2018imagenet} and other semantically irrelevant features \citep{brendel2019approximating, li2018resound}.
Spurious correlations are especially problematic in high-risk domains such as medical imaging, where it was shown that
neural networks can use hospital-specific metal tokens \citep{zech2018variable} or cues of disease treatment \citep{oakden2020hidden} rather than symptoms to
perform automated diagnosis on chest X-ray images.
Spurious features are also extremely prevalent in NLP, where models can achieve good performance on benchmarks without properly solving them,
e.g. by using simple syntactic heuristics such as lexical overlap between the two sentences in order to classify the relationship between them
\citep{niven2019probing, gururangan2018annotation, kaushik2018much, mccoy2019right}.
For a comprehensive survey of the area, see \citet{geirhos2018imagenet}.

Because of the high practical significance of spurious correlations, many \textit{group robustness} methods have been proposed.
These methods aim to reduce the reliance of deep learning models on spurious correlations and improve worst group performance.
Group DRO \citep{sagawa2019distributionally} is the state-of-the-art group robustness method, which minimizes the worst-group loss instead of the average loss.
Other works focus on automatically identifying the minority group examples~\citep{liu2021just, nam2020learning, creager2021environment, zhang2021correct}, learning several diverse classifiers that use different features~\citep{lee2022diversify, pagliardini2022agree,teney2021evading} or using partially available group labels \citep{sohoni2021barack, nam2022spread}.
Group subsampling was shown to be a strong baseline for some benchmarks \citep{idrissi2021simple, sagawa2020investigation}.

In this work, we focus on \emph{feature learning} in the presence of spurious correlations.
\citet{hermann2020shapes} perform a conceptually similar study, but focusing on synthetic datasets.
Similar to their work, we explore how well the different features of the data can be decoded from the features learned by deep neural networks, but on large-scale natural datasets.
\citet{hermann2020origins} explore the feature learning in the context of texture bias \citep{geirhos2018imagenet}, finding that data augmentation has a profound effect on the texture bias while architectures and training objectives have a relatively small effect.
\citet{suvra2022vision} show that Vision Transformer models pretrained on ImageNet22k \citep{kolesnikov2020big} significantly outperform standard CNN models on several spurious correlation benchmarks.

\citet{lovering2020predicting} explore the factors which affect the extractability of features after pre-training and fine-tuning of NLP models.
\citet{Kaushik2020Learning} construct counterfactually augmented sentiment analysis and naural language inference datasets (CAD) and show that combining CAD with the original data reduces the reliance on spurious correlations on the corresponding benchmarks.
\citet{kaushik2021learning} explain the efficacy of CAD and show that while adding noise to causal features degrades in-distribution and out-of-distribution performance, adding noise to non-causal features improves robustess.
\citet{eisenstein2022informativeness} and \citet{veitch2021counterfactual} formally define and study spurious features in NLP from the perspective of causality.

\citet{kirichenko2022last} show that models trained with standard ERM training often learn high-quality representations of the core features, and propose the DFR procedure (see Section~\ref{sec:background}) which we use extensively in this paper.
Related observations have also been reported in other works in the context of spurious correlations \citep{menon2020overparameterisation}, domain generalization~\citep{rosenfeld2022domain} and long-tail classification \citep{kang2019decoupling}.
While we build on the observations of \citet{kirichenko2022last}, our work provides profound new insights and greatly expands on the scope of their work.
In particular, we investigate the feature representations learned by methods beyond standard ERM, and the role of model architecture, pre-training, regularization and data augmentation on learning semantic structure.
We also extend our analysis beyond the standard spurious correlation benchmarks studied by \citet{kirichenko2022last}, 
by considering the challenging real world satellite imaging and chest X-ray datasets.

In an independent and concurrent work, \citet{shi2022robust} also propose an evaluation framework for out-of-distribution generalization based on last layer retraining, inspired by the observations of \citet{kirichenko2022last} and \citet{kang2019decoupling}.
They focus on the comparison of supervised, self-supervised and unsupervised training methods, providing complementary observations to our work.

\vspace{-0.7mm}

\section{Background}
\label{sec:background}

\textbf{Preliminaries.}
\quad
We consider classification tasks with inputs $x \in \mathcal{X}$ and classes $y \in \mathcal{Y}$.
We assume that the data distribution consists of groups $\mathcal{G}$ which are not equally represented in the training data. The distribution of groups can change between the training and test distributions, with majority groups becoming less common or minority groups becoming more common.
Because of the imbalance in training data, models trained with ERM often have a gap between average and worst group performance on test.
Throughout this paper, we will be studying \textit{worst group accuracy} (\textit{WGA}), i.e. the lowest test accuracy across all the groups $\mathcal{G}$. For most problems considered in this paper, we assume that each data point has an attribute $s \in \mathcal{S}$ which is spuriously correlated with the label $y$, and the groups are defined by a combination of the label and spurious attribute: $\mathcal{G} \in \mathcal{Y} \times \mathcal{S}$. 
In test distribution we might find that
$s$ is no longer correlated with $y$, and thus
a model that has learned to rely on the spurious feature $s$ during training will perform poorly at test time.
Models that rely on the spurious features will typically achieve poor worst group accuracy, while models that rely on core features will have more uniform accuracies across the groups.
In Appendix \ref{sec:app_data}, we describe the groups, spurious and core features in the datasets that we use in this paper.

In order to perform controlled experiments, we assume that we have access to the spurious attributes $s$ (or group labels) for training or validation data, which we use for training of group robustness baselines and feature quality evaluation.
However, we emphasize that our results on the features learned by ERM hold generally, even when spurious attributes are unknown, as ERM does not use the information about the spurious features: we only use the spurious attributes to perform analysis.

\textbf{Deep feature reweighting.}
\quad
Suppose we are given a model $m: \mathcal{X} \rightarrow \mathcal{C}$, where  $\mathcal{X}$ is the input space and $\mathcal{C}$ is the set of classes.
\citet{kirichenko2022last} assume that the model $m$ consists of a feature extractor (typically, a sequence of convolutional or transformer layers) followed by a classification head (typically, a single linear layer): $m = h \circ e$, where $e: \mathcal X \rightarrow \mathcal F$ is a feature extractor and $h: \mathcal F \rightarrow \mathcal C$ is a classification head.
They discard the classification head, and use the feature extractor $e$ to compute the set of embeddings $\hat{\mathcal{D}}_e = \{(e(x_i), y_i)\}_{i=1}^n$ of all the datapoints in the reweighting dataset $\hat{\mathcal{D}}$; 
the reweighting dataset is used to retrain the last layer of the model, and contains group-balanced data where the spurious correlation does not hold. 
Finally, they train a logistic regression classifier $l: \mathcal F \rightarrow \mathcal C$ on the dataset $\hat{\mathcal{D}}_e$\footnote{
For stability, logistic regression models are trained $10$ times on different random group-balanced subsets of the reweighting dataset $\hat{\mathcal{D}}$, and the weights of the learned logistic regression models are averaged. See Appendix B of \citet{kirichenko2022last} for full details on the DFR procedure.}.
Then, the final model used on new test data is given by $m_l = l \circ e$.
Thoughout this paper, we use a group-balanced held-out dataset (subset of the validation dataset where each group has the same number of datapoints) as the reweighting dataset $\hat{\mathcal D}$; 
\citet{kirichenko2022last} denote this variation of the method as \DFRVAL \hspace{-1mm}.

\section{Experimental Setup and Evaluation Procedure}
\label{sec:setup}

In this section, we describe the datasets, models and evaluation procedure that we use throughout the paper.

\textbf{Datasets.} \quad
In order to cover a broad range of practical scenarios, we consider four image classification and two text classification problems.

\begin{itemize}

\item \textit{Waterbirds} \citep{sagawa2019distributionally} is a binary image classification problem, where 
the class corresponds to the type of the bird (landbird or waterbird), and the background is spuriously correlated with the class. Namely, most landbirds are shown on land, and most waterbirds are shown over water.

\item \textit{CelebA hair color} \citep{liu2015deep} is a binary image classification problem, where the goal is to predict whether a person shown in the image is blond;
the gender of the person serves as a spurious feature, as $94\%$ of the images with the ``blond'' label depict females.

\item WILDS-\textit{FMOW} \citep{christie2018functional, koh2021wilds, sagawa2021extending} is a satellite image classification problem, where the classes correspond to one of $62$
land use or building types, and the spurious attribute $s$ corresponds to the region (Africa, Americas, Asia, Europe, Oceania or Other; the ``Other'' region is not used in the evaluation).
We note that for the FMOW datasets the groups $\mathcal{G}_s$ are defined by the value of the spurious attribute,
and not the combination of the spurious attribute and the class label $\mathcal {G}_{y, s}$, as described in Section \ref{sec:background}.
Moreover, on FMOW there is also a domain shift: the images for test and validation data (used for last layer retraining) are collected in 2016 and 2017, while the training data is collected before 2016.
For more details, please see Appendix \ref{sec:app_data}.

\item \textit{CXR-14} \citep{wang2017chestx} is a dataset with chest X-ray images for which we focus on a binary classification problem of pneumothorax prediction. \citet{oakden2020hidden} showed that there is a hidden stratification in the dataset such that most images from the positive class contain a chest drain, which is a non-causal feature related to treatment of the disease. While for all other benchmarks we report WGA on test data, for this dataset, following prior work \citep{oakden2020hidden, lee2022diversify, rajpurkar2017chexnet}, we report worst group AUC because of the heavy class imbalance.\footnote{We take the minimum out of the two scores on test: AUC for classifying negative class against positive examples with chest drain and AUC for negative class against positive without chest drain.} 

\item \textit{MultiNLI} \citep{williams2017broad, sagawa2019distributionally} is a text classification problem, where the task is to classify the relationship between a given pair of sentences
as a contradiction, entailment or neither of them.
In this dataset, the presence of negation words (e.g. ``never'') in the second sentence is spuriously correlated with the ``contradiction'' class.

\item \textit{CivilComments} \citep{borkan2019nuanced, koh2021wilds} is a text classification problem, where the goal is to classify whether a given comment is toxic.
We follow \citet{idrissi2021simple} and use the coarse version of the dataset both for training and evaluation, where the spurious attribute is $s = 1$ if the comment mentions at least one of the following categories:
male, female, LGBT, black, white, Christian, Muslim, other religion; otherwise, the spurious label is $0$.
The presence of the eight categories above is spuriously correlated with the comment being classified as toxic.

\end{itemize}

The Waterbirds, CelebA, CivilComments and MultiNLI datasets are commonly used to benchmark the performance of group robustness methods \citep[see e.g.][]{idrissi2021simple, liu2021just, nam2022spread}.
The FMOW and CXR-14 datasets present challenging real-world problems with spurious correlations.
In these datasets, the inputs do not resemble natural images from datasets such as ImageNet~\citep{russakovsky2015imagenet}, 
so models have to learn the relevant features from data to achieve good performance, and cannot simply rely on feature transfer.
We provide detailed descriptions of the data and show example datapoints in Appendix \ref{sec:app_data}, Figures \ref{fig:image_data_desc} and \ref{fig:text_data_desc}.

\textbf{Models.} \quad Following prior work \citep[]{sagawa2019distributionally, idrissi2021simple} we use a ResNet-50 \citep{he2016deep} model pretrained on ImageNet1k \citep{russakovsky2015imagenet} on  Waterbirds, CelebA and FMOW. For the NLP problems, we use a BERT model \citep{devlin2018bert} pre-trained on Book Corpus and English Wikipedia data.
On CXR-14, following prior work \citep[e.g.][]{rajpurkar2017chexnet, oakden2020hidden, lee2022diversify}  we use a DenseNet-121 model~\citep{huang2017densely} pretrained on ImageNet1k.
In Section \ref{sec:architectures}, we provide an extensive study of the effect of architecture and pretraining on the image classification problems,
and in Appendix \ref{sec:app_multnli_res} we perform a similar study on the MultiNLI text classification problem.

\textbf{Evaluation strategy.} \quad
We use DFR 
to evaluate the quality of the learned feature representations, as described in Section \ref{sec:background}: 
we measure how well the core features can be decoded from the learned representations with last layer retraining.
In some of the experiments, we also train 
a linear classifier to predict the spurious attribute $s$ instead of the class label $y$. 
Using this classifier, we can evaluate the decodability of the spurious feature from the learned feature representation.
We refer to this procedure as $s$-DFR and the corresponding worst group accuracy (in predicting the spurious attribute $s$) as \DFRS.
Additionally, we evaluate the worst group accuracy and mean\footnote{
Following \citet{sagawa2019distributionally} and \citet{kirichenko2022last}, we evaluate the mean accuracy according to the group distribution in the training set.
This way, mean accuracy represents the in-distribution generalization of the model.} 
accuracy of the base model without applying DFR, which we refer to as \textit{base WGA} and \textit{base accuracy} respectively.

\section{ERM vs Group Robustness Training}
\label{sec:procedure}

Multiple methods have been proposed for training classifiers which are more robust to spurious correlations, with significant improvements in worst group accuracy compared to standard training.
In this section, 
we use DFR to investigate whether the improvements of group robustness methods are caused by better feature representations or by better weighting of the learned features.

\textbf{Methods.} \quad
We consider 4 methods for learning the features.
\textit{ERM} or Empirical Risk Minimization is the standard training on the original training data, without any techniques targeted at improving worst group performance.
\textit{RWG} reweights the loss on each of the groups according to the size of the group and \textit{RWY} reweights the loss on each class according to the size of the class \citep{idrissi2021simple}.
\textit{Group DRO} \citep{sagawa2019distributionally} is a state-of-the-art method which uses the group information on the training data to minimize the worst group loss instead of the average loss.
Group DRO is often considered as an oracle method or upper-bound on the worst group performance under spurious correlations~\citep{liu2021just, creager2021environment}.

On the CXR dataset the group labels are not available on the train data, so we cannot apply RWG or group DRO; on this dataset we compare ERM to RWY.
On several datasets, the performance of RWY and RWG methods deteriorates during training.
For these datasets, we additionally report the results for the checkpoint obtained with early stopping (RWY-ES and RWG-ES).
For group DRO, we report the performance with early stopping on all datasets except for CXR (GDRO-ES).
In all cases, early stopping is performed based on the worst-group accuracy on the validation set.

\textbf{Hyper-parameters selection.}\quad
We train ERM models, RWG and RWY with the same hyper-parameters shared between all the image datasets (apart from batch size which is set to $32$ on Waterbirds, and $100$ on the other datasets) and between the natural language datasets.
We do not tune the hyper-parameters of these methods for worst group accuracy.
For group DRO, we run a grid search over the values of the generalization adjustment $C$, weight decay and learning rate hyper-parameters, and select the best combination according to the worst-group accuracy on validation data with early stopping.
For details, please see Appendix \ref{sec:app_methods}.

\subsection{Results}

\begin{figure}[t]
\centering
\includegraphics[width=1.05\textwidth]{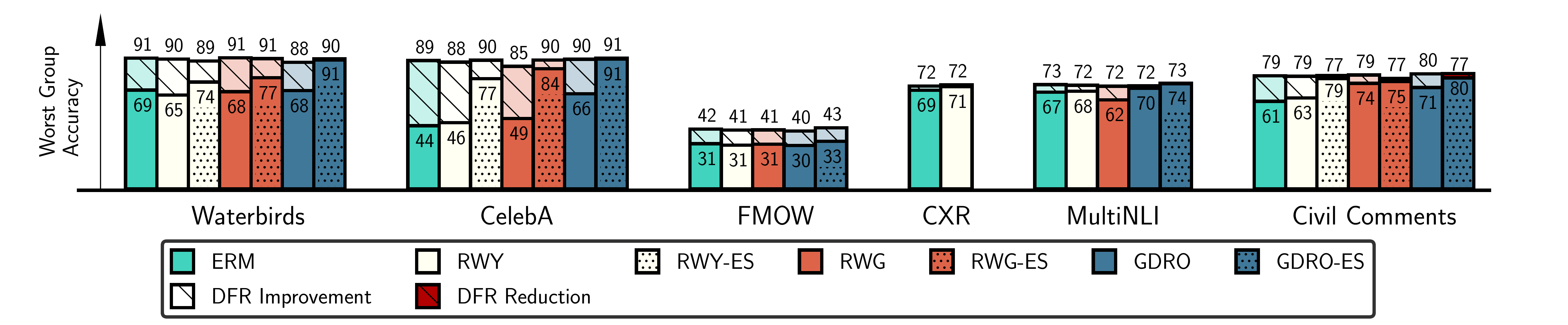}
\caption{
\textbf{ERM vs group robustness methods.}
Performance of  ERM on group robustness methods on vision and NLP benchmark problems.
\textit{ES} stands for early stopping.
For each method on each dataset we report the base model worst group accuracy (shown with the number inside each bar),
and the worst group accuracy after applying DFR (shown above each bar).
On CXR-14, we report worst group AUC.
While on many datasets the base performance of ERM is much worse compared to group robustness methods, the performance
of the different methods is similar after we apply DFR, suggesting that the strong performance of group robustness methods is largely
caused by better weighting of the learned features rather than better feature representations.
}
\label{fig:methods}
\end{figure}

We compare feature learning methods on all datasets in Figure \ref{fig:methods}.
As expected, the worst group accuracy of group robustness methods is significantly better than the ERM worst group accuracy on most datasets.
For example, on Waterbirds, ERM only gets $68.8\%$ WGA, while group DRO with early stopping gets $90.6\%$.
However, after applying DFR the performance of ERM and group DRO is very close, with a slight advantage for ERM ($91.1\%$ for ERM and $90\%$ for group DRO), and similar observations hold on all datasets.

The results for RWG and RWY are analogous. Namely, when combined with early stopping, these methods outperform ERM on base model performance. Once we apply DFR, however, the gap in performance between the methods becomes very small.
In fact, on all the datasets and for all the methods the improvement in worst group accuracy from using any of the considered group robustness methods compared to ERM
does not exceed $1$--$2\%$ after applying DFR.

These results suggest that the improvements over ERM in base model performance for methods such as group DRO and RWG are largely the result of better weighting of the learned features rather than learning better representations of the core features. Indeed, if the core feature was better represented by group robustness methods, we would expect to see a significant improvement over ERM after applying DFR.

This observation is significant both practically and scientifically.
Practically, the problem of training the last layer is simpler, more data efficient and less compute intensive than retraining the full model \citep{kirichenko2022last}.
Our results suggest that for many problems, practitioners can primarily focus on retraining the last layer, as training the feature extractor model with group robustness methods does not provide significant improvements.
Scientifically, robustness to spurious correlations is often implicitly or explicitly associated with the quality of learned feature representations \citep[e.g.][]{arjovsky2019invariant, bahng2020learning, ruan2021optimal, zhang2021correct, lee2022diversify, zhang2022rich}.
Our results suggest that the quality of feature representations, 
i.e. the decodability of the core feature,
is not significantly improved by group DRO, refining our understanding of group robustness training and representation learning in the presence of spurious correlations.

\begin{wrapfigure}{R}{.46\textwidth}
    \vspace{-1.5em}
    \centering
    {\includegraphics[width=1.\linewidth]{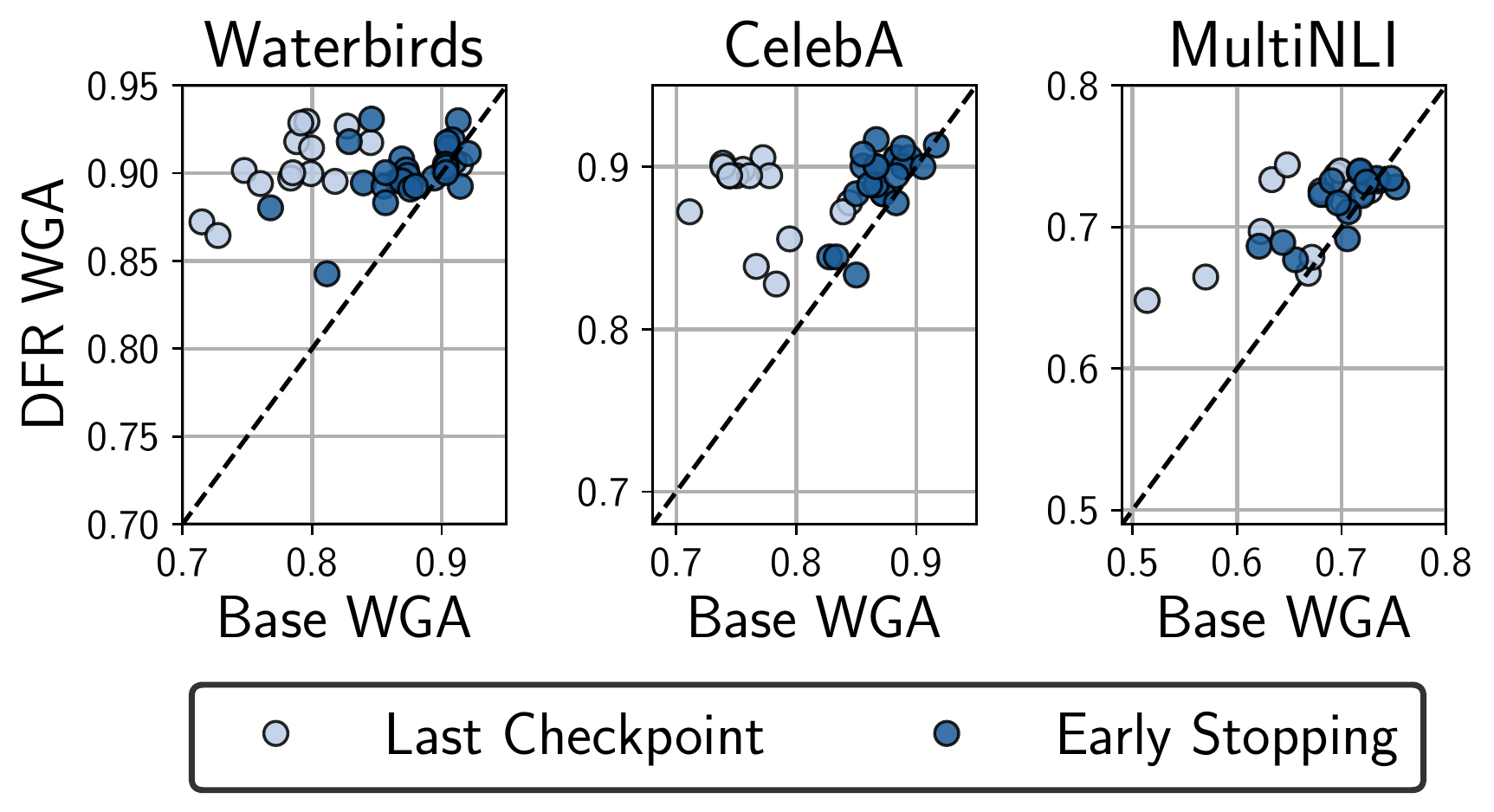}}
    \caption{
    \textbf{Group DRO.} 
    Worst group accuracy before and after applying DFR for a range of group DRO runs.
    DFR does not improve the best runs, indicating that group DRO
    already learns a nearly optimal last layer.
    }
    \label{fig:gdro}
    \vspace{-.5em}
\end{wrapfigure}

\textbf{Effect of early stopping.}\quad
Early stopping is crucial to achieving strong base model performance with RWY, RWG and group DRO on many of the datasets.
In Figure \ref{fig:methods} we report the results both with and without early stopping on datasets where the validation worst group accuracy significantly degrades over the course of training.
Generally, early stopping does not appear to significantly improve the quality of the learned feature representations even in these problems: after applying DFR, methods with and without early stopping achieve similar performance.
This observation suggests that late in training, neural networks may start to assign a higher weight to the spurious features, but the information about the core features is still preserved in the learned representations.
For ERM and Group DRO, we explore the DFR WGA performance as a function of the training iteration in Appendix \ref{sec:app_regularization}, also finding that the length of training has a relatively small effect on the final DFR WGA.

\textbf{Group DRO analysis.}\quad
In Figure \ref{fig:gdro} we report the worst group accuracy of multiple group DRO runs before and after applying DFR.
For each run, we evaluate the best checkpoint according to validation accuracy (i.e. the checkpoint selected by early stopping)
and the last checkpoint saved after a fixed number of epochs.
We observe that while the last checkpoints perform poorly in terms of base WGA in most runs, DFR can significantly improve their performance,
removing the need for early stopping.
Interestingly, we find that the best performing group DRO models cannot be improved by DFR on each of the datasets\footnote{
In Figure \ref{fig:methods} DFR improves the results of GDRO-ES only on the FMOW dataset, where DFR adapts the model to the domain shift, as validation and test images were taken after 2016, while the training images were taken before 2016.
}.
This result suggests that the weighting of the features learned by group DRO is already close to optimal,
again indicating that the success of group DRO can largely be attributed to learning a better weighting for the features in the last linear layer, rather than learning better features.

\begin{mybox}
    Group robustness methods such as group DRO perform well because they improve the last linear layer, not the underlying feature representations.
\end{mybox}

\section{Effect of the Base Model}
\label{sec:architectures}

Most of the prior work on spurious correlation considers a fixed model class for each problem: for example, on Waterbirds and CelebA datasets almost all the papers use a ResNet-50 base model pre-trained on ImageNet1k \citep[e.g.][]{sagawa2019distributionally, liu2021just, lee2022diversify, kirichenko2022last, sohoni2021barack, nam2022spread, zhang2021correct}.
Recently, \citet{suvra2022vision} showed that vision transformer models \citep{dosovitskiy2020image} may provide better robustness to spurious correlations if pretrained on a large dataset.
Here, we perform a systematic large-scale evaluation of the effect of base model choice on the quality of learned feature representations.

We repeat the experiments on the effect of base model, pretraining, and training on the target dataset presented in this section on the MultiNLI text classification task in Appendix \ref{sec:app_multnli_res}, with similar observations.

\subsection{Effect of the architecture and pretraining strategy}

\begin{figure}[t]
\centering
\hspace{-0.cm}\includegraphics[width=0.8\textwidth]{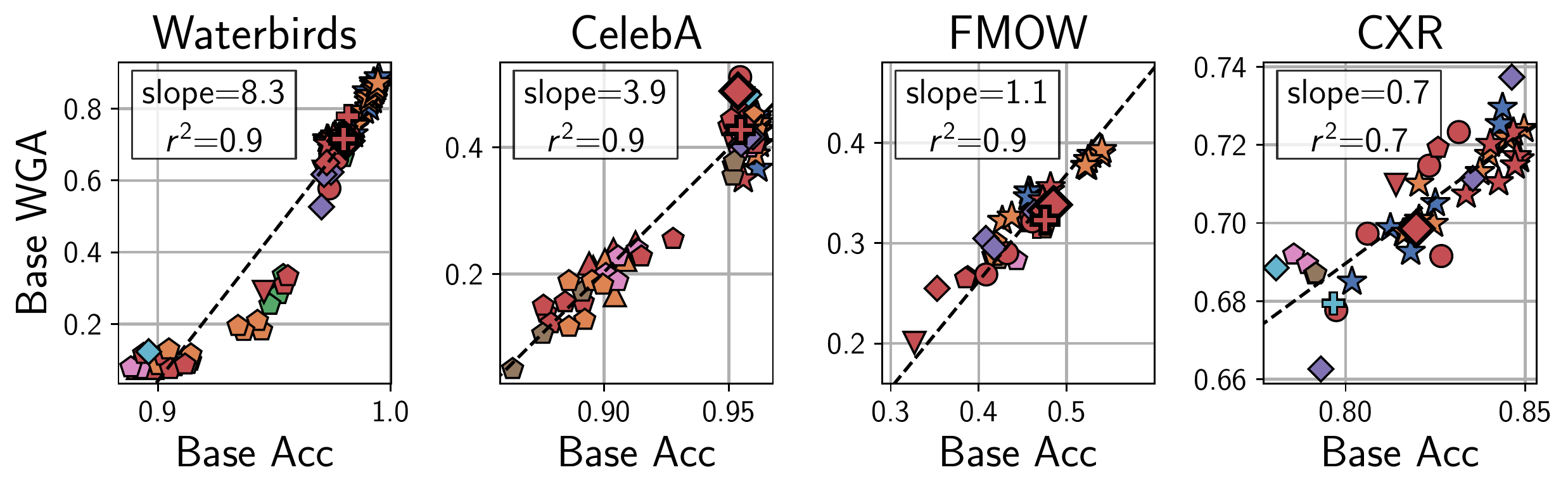} \\
\hspace{-0.cm}\includegraphics[width=0.8\textwidth]{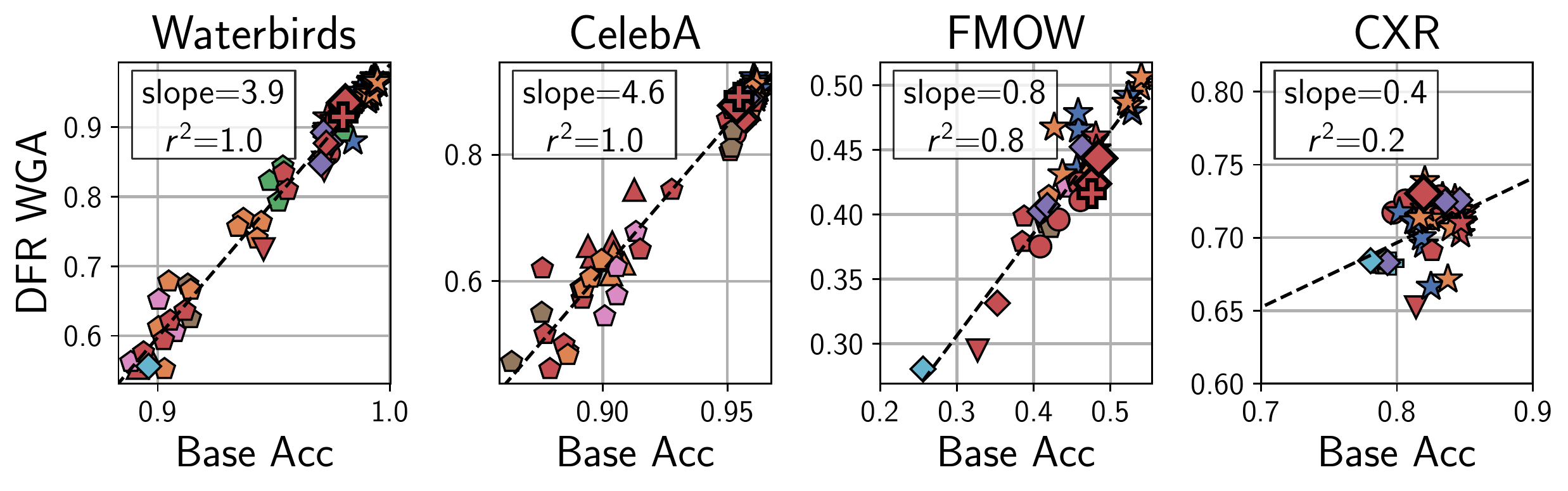}\\[2mm]
\hspace{-0.cm}\includegraphics[trim=0 0 0 10mm, clip, width=.9\textwidth]{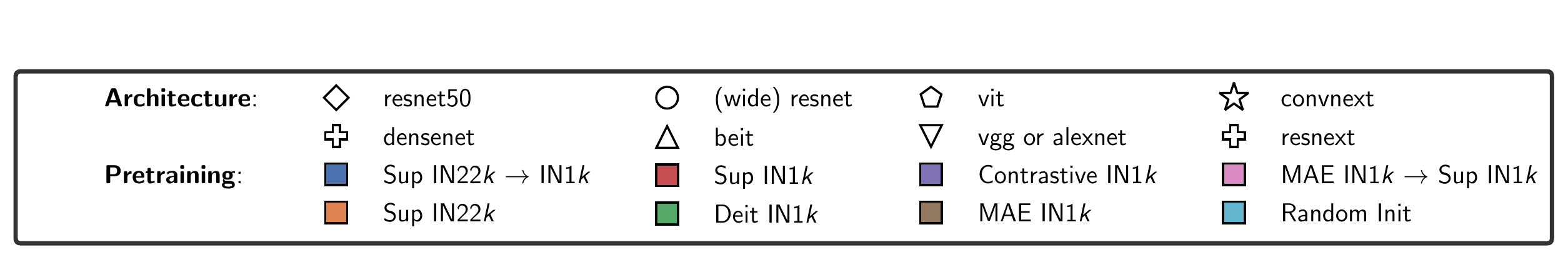}
\caption{
\textbf{Effect of the model architecture and pretraining.}
Base model in-distribution accuracy plotted against base worst-group accuracy (top row) and DFR worst group accuracy (bottom row).
For each panel, we additionally estimate the slope and the $r^2$ score for the linear fit to the data.
On all datasets other than CXR, the DFR WGA is linearly correlated with base model in-distribution performance.
For the base model WGA, the correlation with in-distribution performance is not entirely linear.
Generally, models with better in-distribution performance provide better worst-group performance and better core feature representations.
}
\label{fig:architecture}
\end{figure}

In Figure \ref{fig:architecture}, we plot the base model mean and worst group accuracy and DFR worst group accuracy for a wide range of models and pretraining strategies on each of the four image classification datasets.
We provide model descriptions and training hyper-parameters in Appendix \ref{sec:app_arch_ablation}.
We train a total of 78 models on Waterbirds, 78 on CelebA, 40 on FMOW and 40 on CXR.

\textbf{Accuracy on the line.} \quad
\citet{miller2021accuracy} showed that for many distribution shifts in practice the out-of-distribution performance is highly correlated with the in-distribution generalization performance.
In Figure \ref{fig:architecture} (top row), for each dataset we show scatter plots of base model mean accuracy vs base model worst group accuracy, analogously to \citet{miller2021accuracy}.
For FMOW (which was also considered by \citet{miller2021accuracy}) we observe a linear correlation between the base model mean and worst group accuracies.
On CXR, the dependence is also roughly linear.
However, both on CelebA and on Waterbirds, the correlation does not appear entirely linear.
In particular, on Waterbirds there is a large number of models that have similar worst group accuracy $\approx 20\%$, for which there appears to be little correlation between the base model WGA and mean accuracy.
On CelebA, the same phenomenon occurs for the best performing models, with WGA between $40\%$ and $50\%$.

\textbf{DFR accuracy is on the line.} \quad
Next, in the bottom panels of Figure \ref{fig:architecture}, for each of the datasets we report the base model mean accuracy vs DFR WGA.
On all datasets other than CXR\footnote{
On CXR, we found that the base model is often hard to improve with DFR, suggesting that it already learns to weight the features correctly.
We discuss the possible reasons and the nature of spurious features on this dataset in Appendix \ref{sec:data_discussion}.
}, 
we observe a high linear correlation between the metrics, including the Waterbirds and CelebA.
In particular, the models with the best mean (in-distribution) accuracy also achieve the best DFR WGA.
Note that this is not the case for the base model WGA on CelebA, where the best base model WGA is $51\%$ achieved by
a ResNet-101 model pretrained on ImageNet1k;
this model only achieves mean accuracy of $95.45\%$ compared to 
$96.2\%$ accuracy for the ConvNext XLarge model.
The results in Figure \ref{fig:architecture} confirm that models that achieve better accuracy on the training data distribution generally learn better representations
of the core features, and provide better worst group accuracies with DFR.

The difference between the results for base model WGA and DFR WGA suggests that, analogously to the results in Section \ref{sec:procedure}, some models achieve better worst group performance by choosing a better weighting of the features in the last linear layer, rather than by learning a better representation of the core features.

\textbf{Are VITs more robust than CNNs?}\quad
\citet{suvra2022vision} noted that vision transformers pre-trained on ImageNet22k achieved better worst group accuracies on benchmarks with spurious correlations than popular CNN models.
In particular, with a VIT-B/16 model, they achieve $89.3\%$ worst group accuracy on Waterbirds.
With the ConvNext Large model pretrained on ImageNet22k with ImageNet1k finetuning, we achieve $88.9\%$ worst group accuracy on the Waterbirds dataset.
Notably, ConvNext is a CNN model and \textit{not a vision transformer}.
Generally, we observe that the models that provide the best in-distribution performance also provide better WGA.
In our experiments, we did not observe qualitative differences between the results for vision transformers and CNN models.

\textbf{ERM features are sufficient for SOTA performance.} \quad
The DFR WGA results for the Waterbirds, CelebA and FMOW datasets significantly improve upon the previous best reported results, to the best of our knowledge.
In particular, the ConvNext Large model pretrained on ImageNet22k with ImageNet1k finetuning achieves 97.2\% DFR WGA on Waterbirds and 92.2\% on CelebA;
on FMOW, we only considered smaller ConvNext versions due to computational constraints, still achieving 50.6\% DFR WGA with ConvNext Small pretrained on ImageNet22k; the current best results on the WILDS leaderboard for this dataset are $47.6\%$ followed by $35.5\%$ \citep{koh2021wilds}.
We note that our DFR evaluation uses the validation set to train the last layer of the model, similarly to e.g. \citet{nam2022spread}, and unlike most standard group robustness methods which only use the validation set to tune the parameters.
However, the results presented in this section prove that standard ERM with a strong pretrained model can achieve outstanding results on the robustness benchmarks, significantly improving upon specialized group robustness methods using a weaker model.

\begin{mybox}
    Strong in-distribution generalization correlates with improved robustness to spurious correlations (measured by the DFR worst group accuracy), and this trend holds regardless of the underlying architecture (CNN or ViT).
\end{mybox}

\subsection{Does training on the target data improve features?}

\begin{figure}[t]
\centering
\hspace{-0.cm}\includegraphics[height=0.25\textwidth]{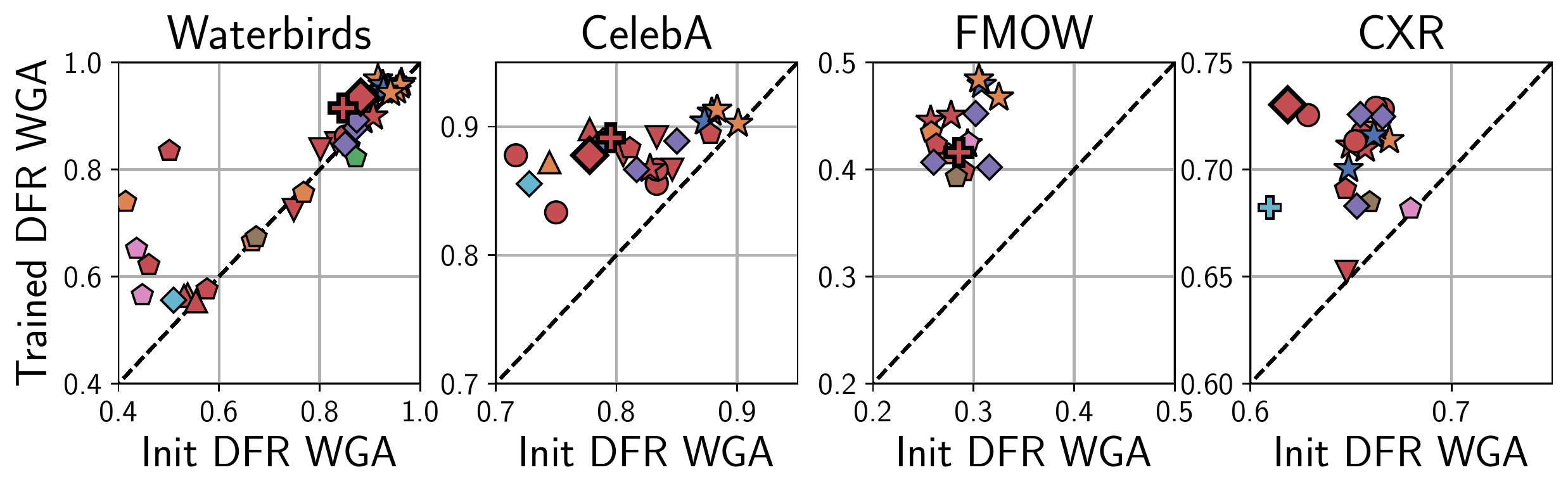} \\
\hspace{-0.cm}\includegraphics[height=0.25\textwidth]{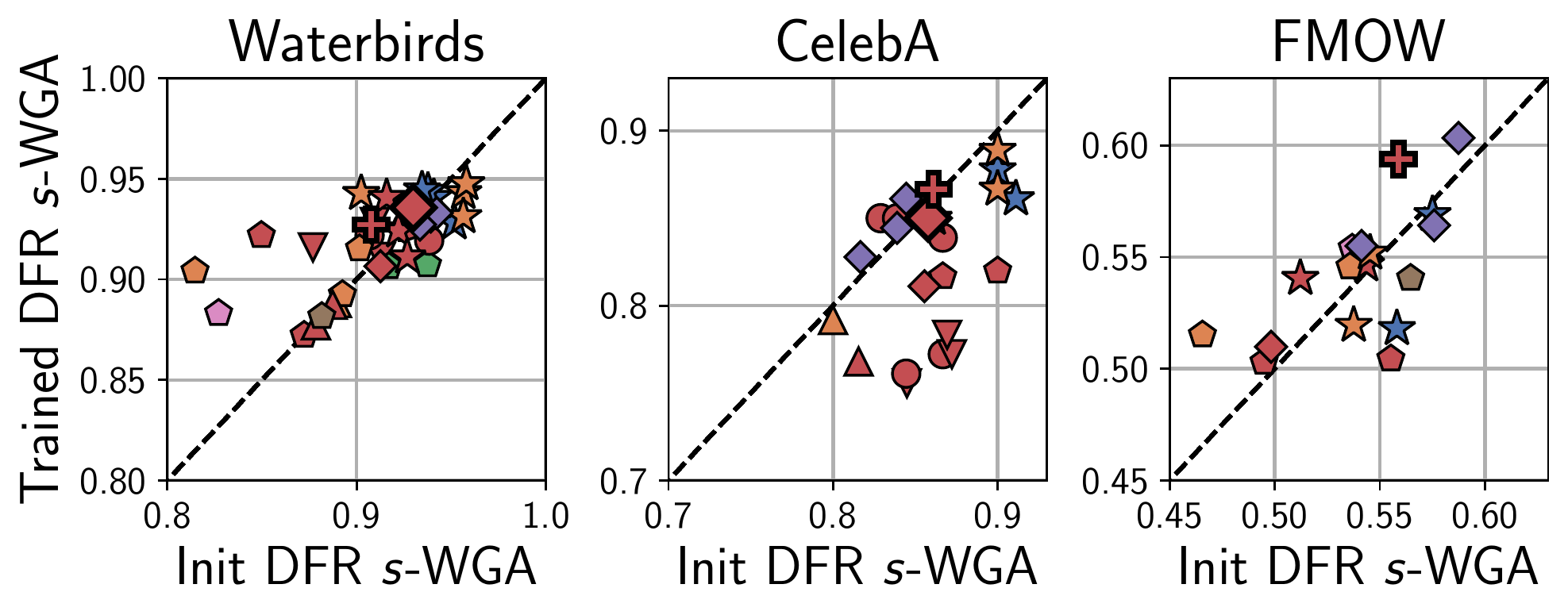} \\[2mm]
\hspace{-0.cm}\includegraphics[trim=0 0 0 10mm, clip, width=.9\textwidth]{figs/arch/legend_2.pdf}
\caption{
\textbf{Effect of training on target data.}
DFR WGA (top 4 panels) and \DFRS~ (bottom 3 panels) before and after training. 
On Waterbirds, the results do not significantly improve from training on the target data.
On the other datasets, the performance improves consistently after training on the target data.
Interestingly, on CelebA the \DFRS~ decreases during training for many models, meaning that the spurious gender feature becomes less predictable from the learned representations.
}
\label{fig:architecture_training}
\end{figure}

Above, we have shown that the choice of the base model architecture and pretraining has a large effect on the quality of the learned feature representations, as measured by DFR WGA.
It is then natural to ask how much of the feature learning happens during training on the target data, and how much is simply transferred from the pretraining task.
To answer this question, for the models trained in the previous section, we run the DFR evaluation on the initial weights of the models, without training the feature extractor on the target data.
We report the \DFR~ and \DFRS~ results in Figure \ref{fig:architecture_training}.
We repeat this experiment on the MultiNLI dataset in Appendix Table \ref{tab:multinli_target}.

Surprisingly, we find that on Waterbirds, for most models the improvement from training on the target (Waterbirds) data is very small, if any.
For example, for the ImageNet1k-pretrained ResNet-50 model that was not trained on Waterbirds data at all\footnote{
This model is used as initialization by most works that report experiments on the Waterbirds dataset.
}, we get $88.2\%$ worst group accuracy by simply training the last layer on the validation data with DFR.
If we finetune the feature extractor on the Waterbirds training data, we can achieve $92.9\%$ DFR WGA.
For reference, the state-of-the-art group DRO method achieves $91\%$ WGA on Waterbirds with this architecture.

Furthermore, with the ConvNext Large model pretrained on ImageNet22k, we get $94\%$ DFR WGA without training the feature extractor on the Waterbirds data,
exceeding the best results previously reported in the literature, to the best of our knowledge. 
This strong performance is not particularly surprising, as ImageNet22k has many of the Waterbirds bird types as classes.
The performance is almost unchanged by training on the target data: DFR WGA after training is $94.3\%$. 
From these results, we can conclude that Waterbirds performance should not be used as a primary metric for feature learning performance, especially if
large-scale pretraining is used!
Indeed, it is possible to achieve outstanding performance, exceeding the previously reported state-of-the-art, without training the feature extractor on the target data.

On the other datasets, the feature learning is more pronounced: the DFR WGA improves after training for all the models considered.
However, on CelebA it is still possible to achieve $88.3\%$ WGA without training on CelebA data,
with ConvNext XLarge pretrained on ImageNet22k, while the best result that we were able to achieve with feature extractors trained on CelebA is $92.2\%$.

\textbf{Is the spurious feature representation improved during training?}\quad
We additionally explore the quality of representation of the spurious feature via \DFRS.
We show the results in Figure~\ref{fig:architecture_training} (bottom row).
Interestingly, on CelebA the spurious gender feature becomes less predictable during training for many of the models.
On FMOW and Waterbirds there is no consistent trend, and the spurious feature does not become significantly more or less predictable during training.

\begin{mybox}
    Strong feature extractors trained on large-scale datasets are sufficient for outstanding performance on benchmark spurious correlation datasets, without any finetuning on the target data.
    Finetuning can improve the quality of the learned representations but often only to a relatively small degree (e.g. on Waterbirds).
\end{mybox}

\subsection{Effect of pretraining strategy}

Finally, using a fixed ResNet-50 model architecture we evaluate the effect of pretraining strategy.
On each of the four datasets, we train a ResNet-50 model initialized with (1)~random initialization, (2) supervised pretraining,
(3) DINO pretraining \citep{caron2021emerging}, (4) SimCLR pretraining \citep{chen2020simple} and (5) Barlow Twins pretraining \citep{zbontar2021barlow} on ImageNet1k.
We report the results in Figure~\ref{fig:pretraining_aug}(a).
On Waterbirds and FMOW datasets, the randomly initialized model does not provide competitive performance.
On CelebA, it still underperforms the pretrained models, but the gap is much smaller.
Among all pretraining methods, supervised is preferable, but the contrastive methods are highly competitive.

In Appendix Table \ref{tab:multinli_pretraining} on MultiNLI with BERT models, we also show that pretraining is crucial for strong performance, while the specific choice of pretraining data has a smaller effect.

\begin{mybox}
    ImageNet pretraining improves the representation of core features on many image datasets with spurious correlations.
\end{mybox}

\section{Effect of Regularization}\label{sec:regularization}

Regularization is used to combat overfitting, including reliance on spurious features.
We consider two regularization techniques: weight decay and data augmentation.

\begin{figure}[t]
\centering
\begin{tabular}{cc}
\hspace{-0.cm}\includegraphics[height=0.23\textwidth]{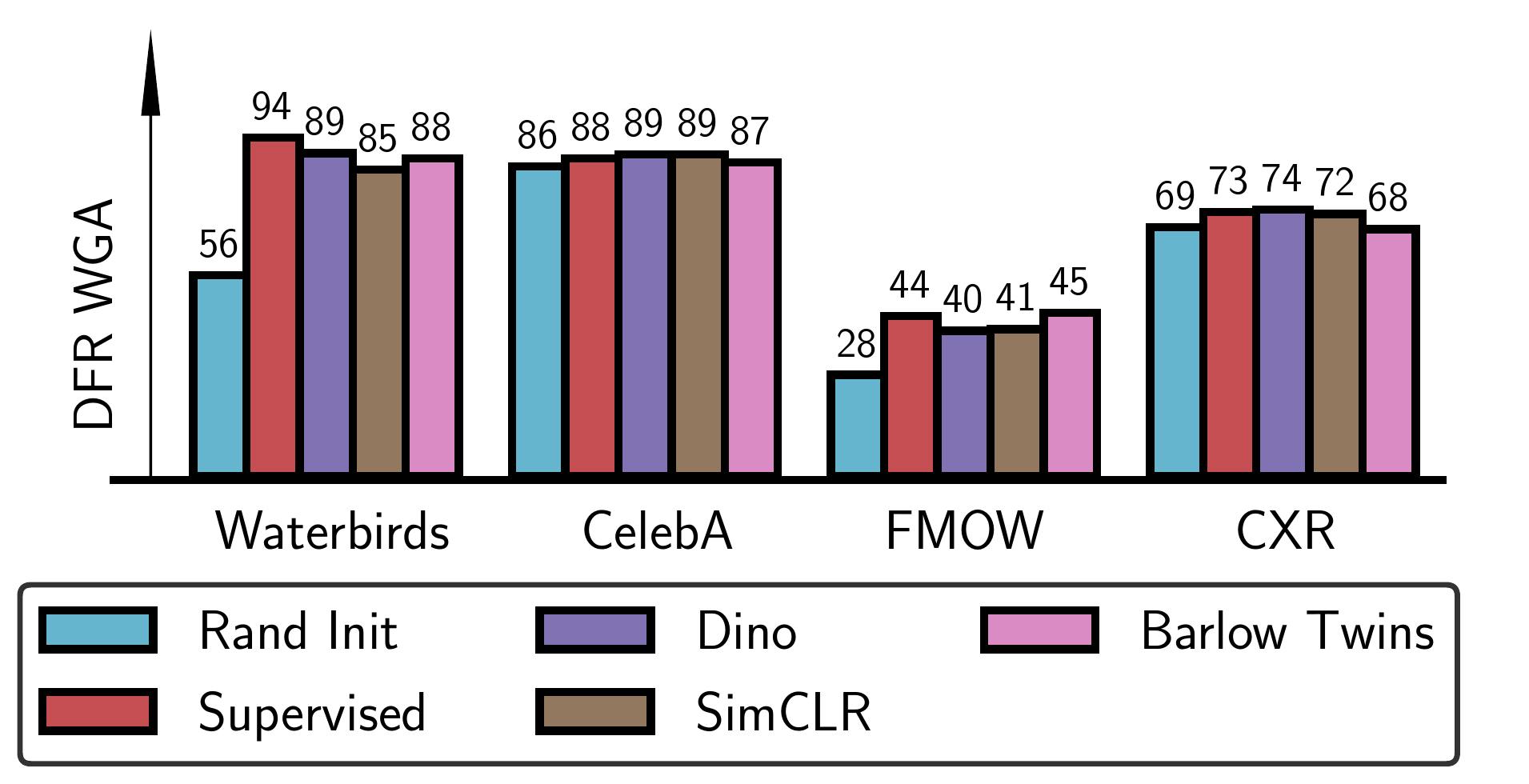} &
\hspace{-0.cm}\includegraphics[height=0.23\textwidth]{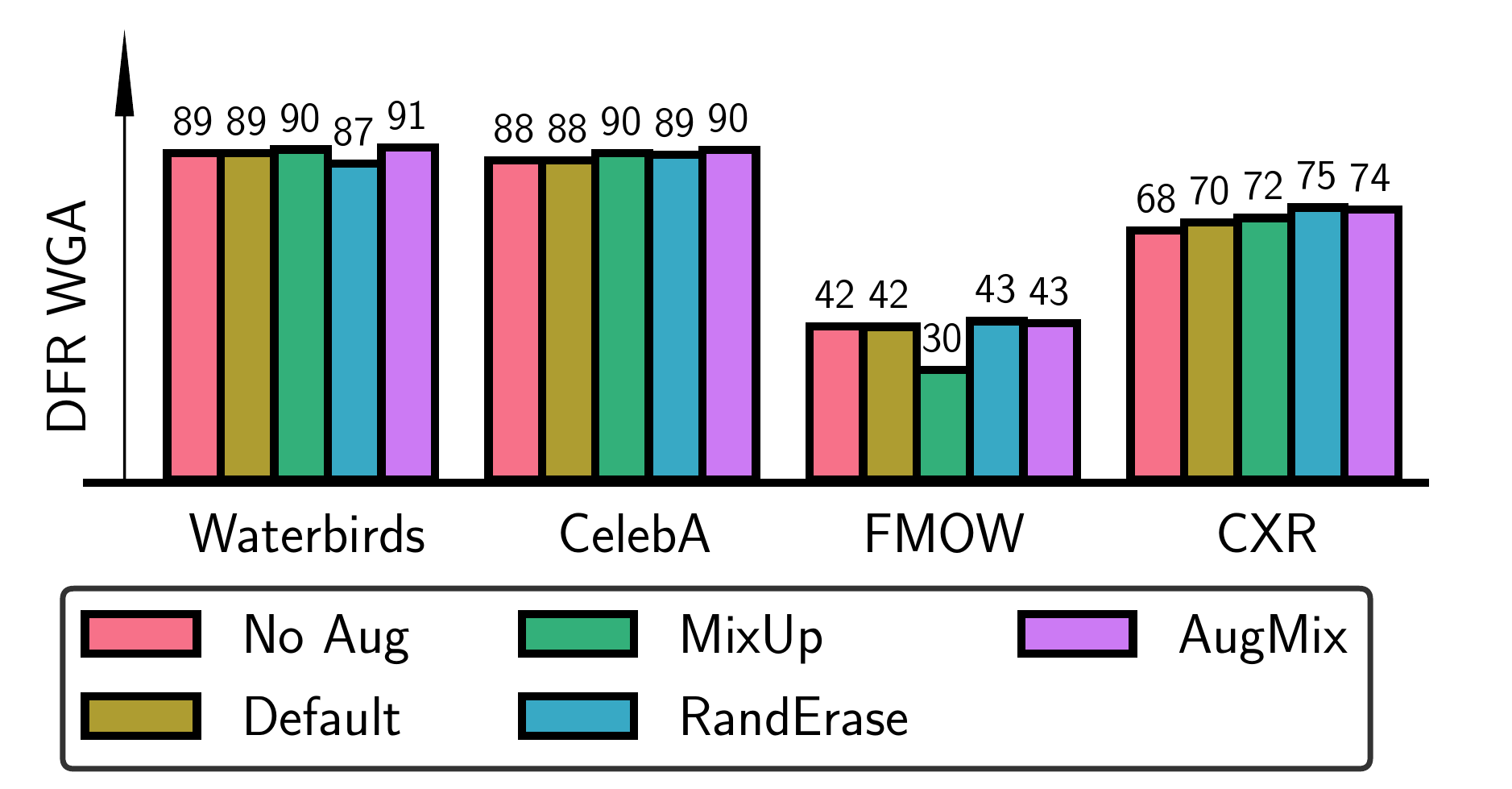}
\\[0.2cm]
\hspace{-0.cm}(a) Pretraining & 
\hspace{-0.cm}(b) Data Augmentation
\end{tabular}
\caption{
\textbf{Effect of pretraining and data augmentation.}
All the results shown use a ResNet50 architecture except for CXR results in panel (b), which use a DenseNet-121 model pretrained on ImageNet.
\textbf{(a)} Supervised pretraining provides the best performance across the board, but contrastive methods are competitive.
Randomly initialized model performs poorly.
\textbf{(b)} AugMix augmentation policy consistently provides strong performance across all datasets, while random erasing and mixup hurt performance on Waterbirds and FMOW respectively.
}
\label{fig:pretraining_aug}
\end{figure}

\begin{wrapfigure}{R}{.44\textwidth}
    \vspace{-.5em}
    \centering
    \includegraphics[width=1.\linewidth]{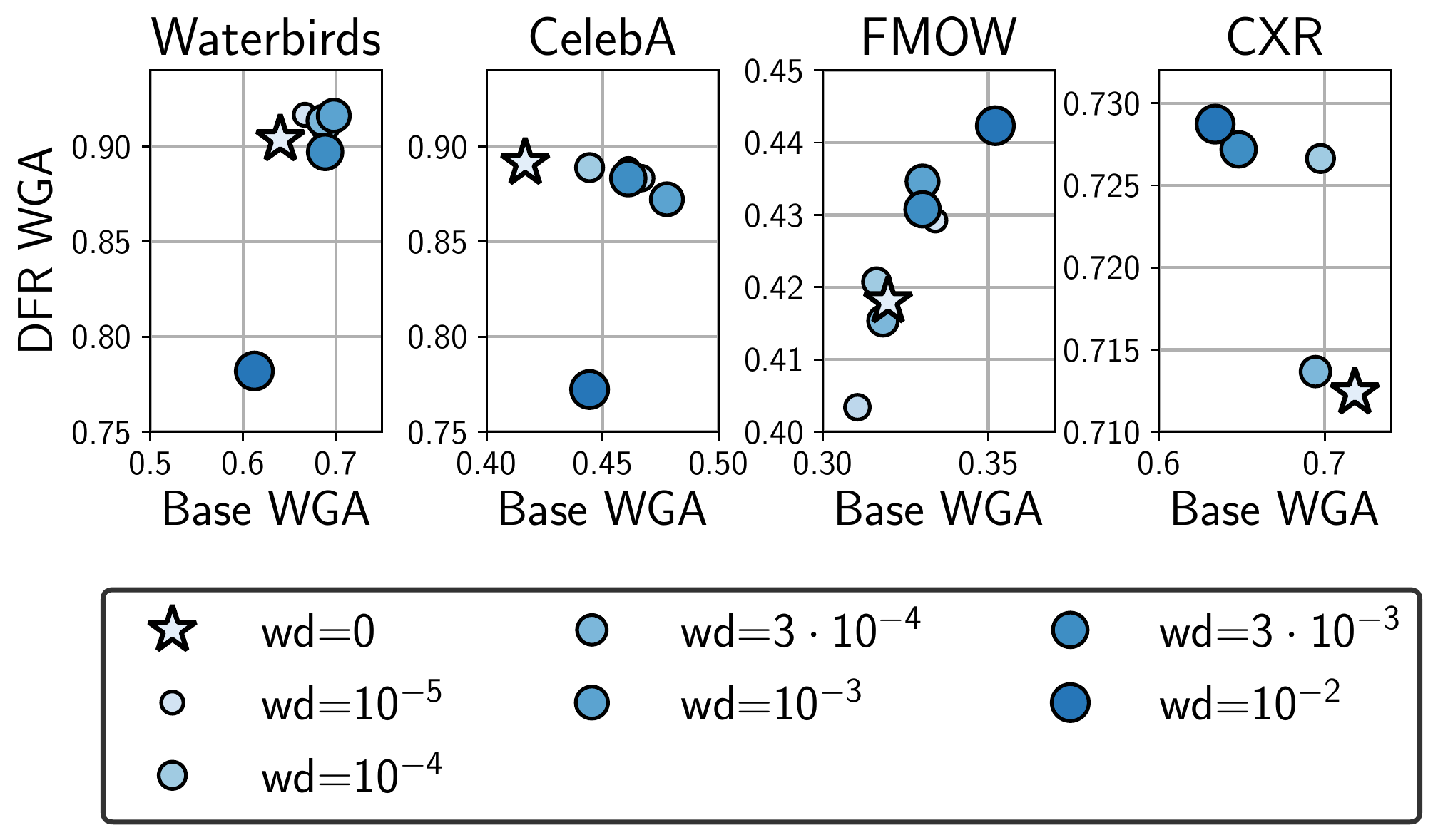}
    \caption{
    \textbf{Effect of weight decay.}
    While zero weight decay underperforms in base WGA on Waterbirds and CelebA, 
    it provides near-optimal DFR WGA.
    }
    \label{fig:wd_effect}
    \vspace{-.5em}
\end{wrapfigure}

\textbf{Effect of weight decay.}\quad
Using a ResNet-50 model pretrained on ImageNet1k, we run training with a range of weight decay values on each of the four datasets, and report the results in Figure \ref{fig:wd_effect}.
For the base model WGA performance, it is generally helpful to set the weight decay to non-zero values.
Indeed, on CelebA the base model WGA for no weight decay is the worst across all weight decay values, losing to the best weight decay by $\approx 5\%$.
However, the no weight decay model is in fact the best model according to DFR WGA!
On these datasets, strong weight decay allows the model to rely less on the spurious features in the last layer (leading to higher base WGA), but does not improve the learned feature representations (similar or worse DFR WGA).
On other datasets no weight decay also provides competitive performance.
FMOW is the only dataset considered where non-zero weight appears to be helpful for learning high quality representations of the core features.
In Appendix Table \ref{tab:multinli_wd} we report analogous results on the MultiNLI text classification problem.

Generally, weight decay affects both the feature extractor and the last classification layer.
While in some cases higher weight decay is helpful for learning better features, zero weight decay is competitive across the board.
This observation is especially interesrting given that \citet{sagawa2019distributionally} showed that group DRO requires stronger than usual weight decay to achieve good performance on Waterbirds and CelebA datasets.

\textbf{Effect of Data Augmentation.}\quad
Next, we consider 5 data augmentation policies: (1)~no augmentations, (2) default augmentations, i.e. random crops and horizontal flips, (3)~MixUp~\citep{zhang2017mixup} combined with default augmentations, (4) Random Erasing \citep{zhong2020random}, and (5)~AugMix \citep{hendrycks2019augmix}.
We train ImageNet1k-pretrained DenseNet-121 models on CXR and ResNet-50 models on Waterbirds, CelebA and FMOW with each of the augmentation policies,
and report the results in Figure \ref{fig:pretraining_aug}(b).
AugMix provides the best performance on each dataset.
However, we find that data augmentation is generally not required to achieve strong performance on any of the datasets with the exception of CXR:
the model trained without augmentations is competitive across the board.
Moreover, data augmentation can \textit{hurt} the learned features.
For example, while MixUp is helpful on Waterbirds and CelebA, it significantly hurts the preformance on FMOW, with $30\%$ DFR WGA compared to $42\%$ for the model trained without augmentation.
We hypothesize that on the FMOW dataset, which contains highly detailed satellite images, mixing the images makes training overly challenging.

Similarly, Random Erasing hurts the DFR WGA on Waterbirds.
We hypothesize that the randomly erased image block is more likely to fully cover the bird features than the background features, as the bird occupies a small fraction of the image relative to the background.
Consequently, the model is incentivised to focus on the spurious feature:
the model trained with Random Erasing learned the highest quality representation of the spurious feature across all augmentations with \DFRS~of $91.8 \pm 0.1\%$ across $3$ runs,
compared to e.g. $91.4 \pm 0.5\%$ for the model trained with no augmentation.

\citet{balestriero2022effects} report a related observation for models trained on ImageNet: in some cases data augmentation affects different classes disproportionately, and best data augmentation policies for mean accuracy can lead to poor worst-class accuracy.

\begin{mybox}
    Strong regularization is not necessary for robustness to spurious correlations, but appropriate data augmentation and weight decay can provide a small improvement.
\end{mybox}

\section{Discussion}

The worst group performance of a model is affected by two factors: the quality of the representation of the core features produced by the feature extractor and the weight assigned to the core features in the last classification layer. 
In contrast to prior work, we consider the quality of the feature extractor in isolation, focusing on realistic datasets and large-scale models.
We find that many of the popular group robustness methods improve the worst group performance primarily by learning a better last layer and not by learning a better feature representation.
Similarly, regularization techniques such as early stopping and strong weight decay can improve the worst group accuracy by learning a better last layer, but do not lead to a consistent improvement in terms of the quality of the learned feature representations.
On the other hand, the base model architecture and pre-training strategy have a major effect on the quality of the feature representations.

Our observations suggest an important open question: is it possible to significantly improve upon standard ERM in terms of the quality of the learned representations for a given base model?
In future work, it would be interesting to evaluate methods such as Rich Feature Construction~\citep{zhang2022rich}, gradient starvation \citep{pezeshki2021gradient}, ensembling \citep{lakshminarayanan2017simple} and other techniques for increasing feature diversity~\citep[e.g.][]{lee2022diversify}.
We hope that our work will also inspire new group robustness methods targeted specifically at improving the quality of the core feature representations.

{ \small
\bibliographystyle{apalike}
\bibliography{neurips}
}

\newpage

\appendix

\section*{Appendix Outline}

This appendix is structured as follows.
In Section \ref{sec:app_data} we describe the datasets, augmentation policies and models used in this paper.
In Section \ref{sec:app_methods} we provide details on the methods, implementations and hyper-parameters, as well as detailed results for the experiments in Section \ref{sec:procedure}.
In Section \ref{sec:app_arch_ablation} we provide additional details on the experiments in Section \ref{sec:architectures}.
In Section \ref{sec:app_regularization} we provide additional details and results for the experiments in Section \ref{sec:regularization}.
In Section \ref{sec:app_multnli_res} we provide additional results on the MultiNLI dataset.
Finally, in Section \ref{sec:limitations} we describe the limitation, broader impact, compute and licenses.

\textbf{Tools and packages.}\quad
During the work on this paper, we used the following tools and packages:
\lstinline{NumPy} \citep{harris2020array}, \lstinline{SciPy}~\citep{2020SciPy-NMeth}, \lstinline{PyTorch} \citep{paszke2017automatic}, \lstinline{TorchVision} \citep{marcel2010torchvision}, \lstinline{Jupyter notebooks} \citep{Kluyver2016jupyter}, \lstinline{Matplotlib} \citep{Hunter:2007}, \lstinline{Pandas} \citep{pandas}, \lstinline{Weights&Biases} \citep{wandb},
\lstinline{timm} \citep{rw2019timm}, \lstinline{transformers} \citep{wolf2019huggingface}, \lstinline{vissl} \citep{goyal2021vissl}.

\section{Data and Models}
\label{sec:app_data}

In this section, we describe the datasets, data augmentation policies and models used throughout the paper.

\subsection{Datasets}
\begin{figure}[h]
\centering

\vspace{-2cm}
\fboxother{
\begin{tabular}{cc cccc}

\textbf{Waterbirds} &
\multicolumn{5}{r}{
\textbf{Target}: bird type;\quad \textbf{Spurious feature}: background type.} \\
\hline\\[-3mm]

\textbf{Image:} && 
\begin{tabular}{c}\includegraphics[width=0.15\textwidth]{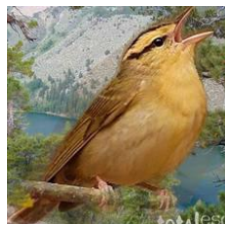}\end{tabular} &
\begin{tabular}{c}\includegraphics[width=0.15\textwidth]{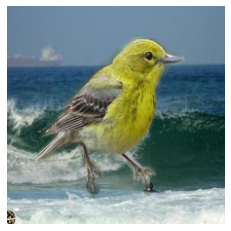}\end{tabular} &
\begin{tabular}{c}\includegraphics[width=0.15\textwidth]{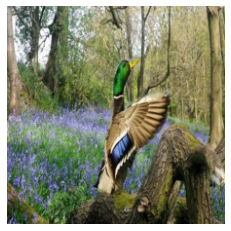}\end{tabular} &
\begin{tabular}{c}\includegraphics[width=0.15\textwidth]{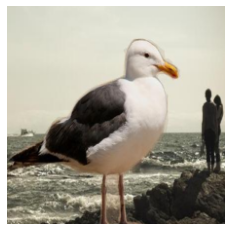}\end{tabular}
\\
\rowcolor{rowcolor}
\textbf{Group $g$:} &\quad\quad& 
$0$
& $1$
& $2$
& $3$ 
\\
\textbf{Target $y$:}    && 0 & 0 & 1 & 1
\\
\rowcolor{rowcolor}
\textbf{Spurious $s$:}    && 0 & 1 & 0 & 1
\\
\textbf{Description:} &&
\begin{tabular}{c} landbird \\[-1mm] on land \end{tabular} &
\begin{tabular}{c} landbird \\[-1mm] on water \end{tabular} &
\begin{tabular}{c} waterbird \\[-1mm] on land \end{tabular} &
\begin{tabular}{c} waterbird \\[-1mm] on water \end{tabular}
\\
\rowcolor{rowcolor}
\textbf{\# Train data:}  &&
3498 (73\%) &  184 (4\%) & 56 (1\%) & 1057 (22\%)
\\
\textbf{\# Val data:}    &&
467 & 466 & 133 & 133
\end{tabular}
}
\\[5mm]

\fboxother{
\begin{tabular}{cc cccc}

\multicolumn{3}{l}{\textbf{CelebA hair color}} &
\multicolumn{3}{r}{
\textbf{Target}: hair color;\quad\quad \textbf{Spurious feature}: gender.} \\
\hline\\[-3mm]

\textbf{Image:} && 
\begin{tabular}{c}\includegraphics[width=0.15\textwidth]{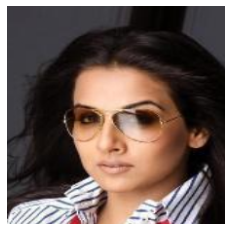}\end{tabular} &
\begin{tabular}{c}\includegraphics[width=0.15\textwidth]{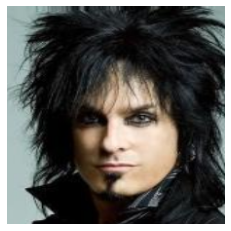}\end{tabular} &
\begin{tabular}{c}\includegraphics[width=0.15\textwidth]{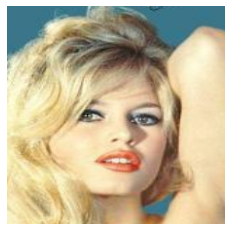}\end{tabular} &
\begin{tabular}{c}\includegraphics[width=0.15\textwidth]{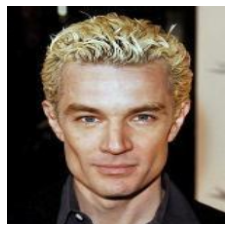}\end{tabular}
\\
\rowcolor{rowcolor}
\textbf{Group $g$:} &\quad\quad& 
$0$
& $1$
& $2$
& $3$ 
\\
\textbf{Target $y$:}    && 0 & 0 & 1 & 1
\\
\rowcolor{rowcolor}
\textbf{Spurious $s$:}    && 0 & 1 & 0 & 1
\\
 \textbf{Description:} &&
\begin{tabular}{c} non-blond \\[-1mm] woman \end{tabular} &
\begin{tabular}{c} non-blond \\[-1mm] man \end{tabular} &
\begin{tabular}{c} blond \\[-1mm] woman \end{tabular} &
\begin{tabular}{c} blond \\[-1mm] man \end{tabular}
\\
\rowcolor{rowcolor}
\textbf{\# Train data:} &&
71629 (44\%) & 66874 (41\%) & 22880 (14\%) &  1387 (1\%)
\\
\textbf{\# Val data:} &&
8535 & 8276 & 2874 &  182
\end{tabular}
}
\\[5mm]

\fboxother{
\begin{tabular}{c ccccc}

\multicolumn{2}{l}{\textbf{Wilds-FMOW}} &
\multicolumn{4}{r}{
\textbf{Target}: land use / building;\quad\quad \textbf{Spurious feature}: region.} \\
\hline\\[-3mm]

\textbf{Image:} &~~
\begin{tabular}{c}\includegraphics[width=0.115\textwidth]{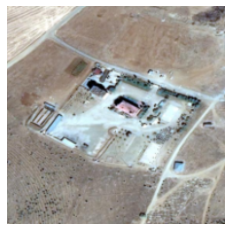}\end{tabular} &
\begin{tabular}{c}\includegraphics[width=0.115\textwidth]{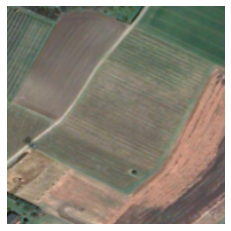}\end{tabular} &
\begin{tabular}{c}\includegraphics[width=0.115\textwidth]{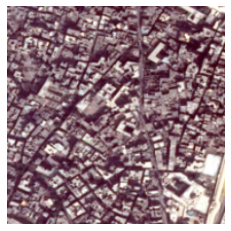}\end{tabular} &
\begin{tabular}{c}\includegraphics[width=0.115\textwidth]{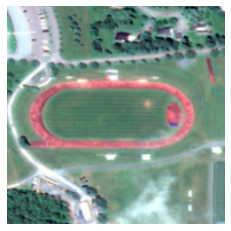}\end{tabular}
&
\begin{tabular}{c}\includegraphics[width=0.115\textwidth]{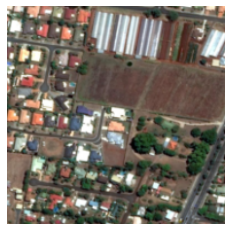}\end{tabular}
\\
\rowcolor{rowcolor}
\textbf{Group $g$:} &
$0$
& $1$
& $2$
& $3$
& $4$ 
\\
\textbf{Target $y$:}    & \{0,\ldots,61\} & \{0,\ldots,61\} & \{0,\ldots,61\} & \{0,\ldots,61\} & \{0,\ldots,61\}
\\
\rowcolor{rowcolor}
\textbf{Spurious $s$:}    & 0 & 1 & 2 & 3 & 4
\\
 \textbf{Description:} &
Asia & Europe & Africa & America & Oceania
\\
\rowcolor{rowcolor}
\textbf{\# Train data:} &
17809 (23\%) & 34816 (45\%) & 1582 (2\%) &  20973 (27\%) & 1641 (2\%)
\\
\textbf{\# Val data:} &
4121 & 7732 & 803 &  6562 & 693
\end{tabular}
}
\\[5mm]

\fboxother{
\begin{tabular}{cc ccc}

\multicolumn{2}{l}{\textbf{CXR-14}} &
\multicolumn{3}{r}{
\textbf{Target}: pneumothorax;\quad\quad \textbf{Shortcut feature}: chest drain.} \\
\hline\\[-3mm]

\textbf{Image:} && 
\begin{tabular}{c}\includegraphics[width=0.19\textwidth]{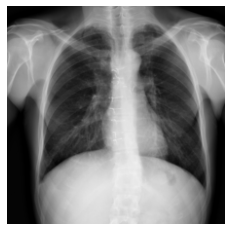}\end{tabular} 
\quad\quad&
\begin{tabular}{c}\includegraphics[width=0.19\textwidth]{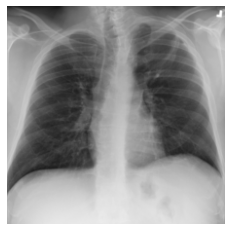}\end{tabular}
\quad\quad&
\begin{tabular}{c}\includegraphics[width=0.19\textwidth]{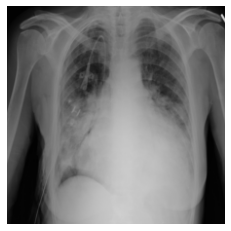}\end{tabular}
\\
\rowcolor{rowcolor}
\textbf{Group $g$:} &\quad\quad~& 
$0$
& $1$
& $2$
\\
\textbf{Target $y$:}    && 0 & 1 & 1
\\
\rowcolor{rowcolor}
\textbf{Spurious $s$:}    && 0 & 0 & 1
\\
\textbf{Description:} &&
not sick, no chest drain &
sick, no chest drain  &
sick, chest drain
\\
\rowcolor{rowcolor}
\textbf{\# Train data:} &&
71629 (95\%) & ? &  ?
\\
\textbf{\# Val data:} &&
10714 & 204 & 300
\\
\end{tabular}
}
\\[3mm]

\caption{
\textbf{Image datasets.}
Group descriptions and example images for Waterbirds, CelebA hair color, Wilds-FMOW and CXR-14 datasets.
Each column corresponds to a group in the dataset.
For CXR-14 group labels are not known on the training data.
}
\label{fig:image_data_desc}
\end{figure}

We perform experiments on $4$ image classification and $2$ text classification problems.
We illustrate the image datasets in Figure \ref{fig:image_data_desc} and the text datasets in Figure \ref{fig:text_data_desc}.

\textbf{Waterbirds.}\quad
The Waterbirds dataset is described in Figure \ref{fig:image_data_desc}.
The dataset contains images of birds from the CUB dataset \citep{wah2011caltech} pasted on the backgrounds from the Places dataset \citep{zhou2017places}.
The spurious attribute $s$ describes the type of background (water or land) and the core feature associated with the target $y$ is the type of the bird (waterbird or landbird).
For a detailed description of the data generating process, see \citet{sagawa2019distributionally}.
The background is spuriously correlated with the bird type in the training datasets: waterbirds are more likely to be placed on a water background, and landbirds are more likely to be placed on a land background.
There are $4$ groups defined to the tuples $(y, s)$.

\textbf{CelebA.}\quad
The CelebA hair color dataset is described in Figure \ref{fig:image_data_desc}.
The dataset contains photos of celebrities from the CelebA dataset \citet{liu2015deep}.
The core attribute associated with the target $y$ is the hair color (blond vs non-blond).
The gender serves as a spurious feature $s$: the vast majority of blond people in CelebA are female.
There are $4$ groups defined to the tuples $(y, s)$.

\textbf{FMOW.}\quad
The WILDS-FMOW dataset is described in Figure \ref{fig:image_data_desc}.
This dataset is a part of the WILDS benchmark \citep{koh2021wilds, sagawa2021extending}, and was originally collected by \citet{christie2018functional}.
The dataset contains satellite images, and the target $y$ describes the type of building or land use shown in the image.
There are $62$ classes.
The spurious attribute $s$ corresponds to the region (Asia, Europe, Africa, America, Oceania) shown in the image.
The training data additionally contains another group \textit{Other}, which is dropped during evaluation.
For this dataset, following the WILDS benchmark, we define groups by just the value of the spurious attribute: $g = s$.
In particular, worst group accuracy corresponds to the worst accuracy across regions.
The regions are represented unequally in the data, leading to unequal performance.
Moreover, the test images were taken several years later than the train images, constituting an additional type of distribution shift.
We note that for the DFR evaluation, we use the validation data to train the last layer of the model, which means that our results would be classified as non-standard submissions to the WILDS leaderboard.

\FloatBarrier

\begin{figure}[t]
\footnotesize
\centering
\resizebox{\textwidth}{!}{
\fboxother{
\begin{tabular}{cccccc}

\multicolumn{1}{l}{\textbf{Civil Comments}} &
\multicolumn{5}{r}{
\textbf{Target}: toxic / neutral comment;} \\
\multicolumn{1}{l}{} &
\multicolumn{5}{r}{
\textbf{Spurious feature}: mentions protected categories.} \\
\hline\\[-3mm]

\textbf{Example} & \textbf{Group} $g$ & \textbf{Target $y$} & \textbf{Spur. $s$} & 
\begin{tabular}{c}\textbf{\# Train} \\ \textbf{data} \end{tabular} &
\begin{tabular}{c}\textbf{\# Val} \\ \textbf{data} \end{tabular}
\\[4mm]
\rowcolor{rowcolor}
\begin{tabular}{c}
\small{\textit{``I wouldn't think this would be so rare on the}} \\
\small{\textit{plains of eastern Colorado.''}}
\end{tabular}&
0 & 0 (Netral) & 0 & 148186 (55\%) & 25159 
\\
\begin{tabular}{c}
\small{\textit{``If the person wanted to write to the Bishop, he or she would have. }} \\
\small{\textit{They wanted the Vatican to know their pain, not to acknowledge it}} \\
\small{\textit{is a lack of Christian charity and kindness.''}}
\end{tabular}&
1 & 0 (Netral) & 1 & 90337 (33\%) & 14966 \\
\rowcolor{rowcolor}
\begin{tabular}{c}
\small{\textit{``What a gross example of bureaucrats and lawyers showing  }} \\
\small{\textit{everyone in Oregon who our bosses are. Next the jerks will}} \\
\small{\textit{demand we bow to them. <...>''}}
\end{tabular}&
2 & 1 (Toxic) & 0 & 12731 (5\%) & 2111 \\
\begin{tabular}{c}
\small{\textit{``Democrats, RINO's and atheists won't be happy until they }} \\
\small{\textit{have destroyed conservatives and christians.''}}
\end{tabular} &
3 & 1 (Toxic) & 1 & 17784 (7\%) & 2944 \\
\end{tabular}
}
}
\\[5mm]

\resizebox{\textwidth}{!}{
\fboxother{
\begin{tabular}{cccccc}
\multicolumn{1}{l}{\textbf{MultiNLI}} &
\multicolumn{5}{r}{
\textbf{Target}: contradiction / entailment / neutral;} \\
\multicolumn{1}{l}{} &
\multicolumn{5}{r}{
\textbf{Spurious feature}: has negation words.} \\
\hline\\[-3mm]

\textbf{Example} & \textbf{Group} $g$ & \textbf{Target $y$} & \textbf{Spur. $s$} & 
\begin{tabular}{c}\textbf{\# Train} \\ \textbf{data} \end{tabular} &
\begin{tabular}{c}\textbf{\# Val} \\ \textbf{data} \end{tabular}
\\[4mm]

\rowcolor{rowcolor}
\small{\textit{``he was up quickly. [SEP] he sat the entire time and didn't move.''}} &
0 & 0 (contr.) & 0 & 57498 (28\%) & 22814
\\
\begin{tabular}{c}
\small{\textit{``for golf enthusiasts, two courses reassure the visitor }} \\
\small{\textit{you are close to civilization. [SEP] there is no golf }} \\
\small{\textit{ course available anywhere around.''}}
\end{tabular}&
1 & 0 (contr) & 1 & 11158 (5\%) & 4634 \\
\rowcolor{rowcolor}
\begin{tabular}{c}
\small{\textit{``while emergency physicians may not have the time or interest, }} \\
\small{\textit{the patients do. [SEP] the patients have time and interest, }} \\
\small{\textit{unlike emergency physicians.''}}
\end{tabular}&
2 & 1 (entail.) & 0 & 67376 (32\%) & 26949 \\
\begin{tabular}{c}
\small{\textit{``then it dawned on him that of course the lawyer did not know. }} \\
\small{\textit{[SEP] he realised that the lawyer had no idea.''}}
\end{tabular} &
3 & 1 (entail.) & 1 & 1521 (1\%) & 613 \\
\rowcolor{rowcolor}
\begin{tabular}{c}
\small{\textit{``disneyland is huge and can be very crowded in summer. }} \\
\small{\textit{[SEP] going to disneyland is every child's dream.}}
\end{tabular}&
4 & 2 (neutr.) & 0 & 66630 (32\%) & 26655 \\
\begin{tabular}{c}
\small{\textit{``you have raced him, senor? " he asked drew with formal }} \\
\small{\textit{courtesy. [SEP] drew replied that he had never raced him.''}}
\end{tabular} &
5 & 2 (neutr.) & 1 & 1992 (1\%) & 797 \\

\end{tabular}
}}
\\[5mm]

\caption{
\textbf{Text datasets.}
Text examples, class labels, spurious attributes and group labels for the Civil Comments and MultiNLI datasets.
On both datasets, the spurious feature is correlated with the class label.
}
\label{fig:text_data_desc}
\end{figure}

\textbf{CXR.}\quad
The CXR dataset is described in Figure \ref{fig:image_data_desc}.
The images are taken from the CXR-14 dataset \citep{wang2017chestx}.
CXR-14 is a multi-label dataset, where each label corresponds to a disease, and one image can show multiple diseases.
We perform the pneumothorax classification task, i.e. all images that have the pneumothorax label have $y=1$ and all images that do not have $y=0$.
The dataset contains several images for some of the patients; there is no patient overlap between the train, test and validation splits.
\citet{oakden2020hidden} identified a hidden stratification in this dataset: a lot of images showing patients with pneumothorax showed a \textit{chest drain}, which
is a treatment for the pneumothorax disease.
The neural networks trained on this dataset are using the chest drain as a shortcut feature, and perform much worse when classifying images without the chest drain.
The labels for the spurious feature are only available on the validation and test datasets, and only for the images showing sick patients, i.e. there is no $(y=0, s=1)$ group.
There are $3$ groups corresponding to available pairs $(y, s)$.
Following prior work \citep[e.g.][]{oakden2020hidden,lee2022diversify,rajpurkar2017chexnet}, we compute three AUC values: for classifying group $0$ against group $1$, group $0$ against group $2$ and group $0$ against the combined groups $0$ and $1$;
instead of worst group accuracy we report the lower of the first two AUC values, and instead of mean accuracy we report the last AUC value throughout the experiments.

\textbf{Civil Comments}\quad
The Civil Comments Coarse dataset is described in Figure \ref{fig:text_data_desc}.
The dataset was originally collected in \citet{borkan2019nuanced} and is a part of the WILDS benchmark \citep{koh2021wilds, sagawa2021extending}.
This dataset contains comments that are classified as toxic or not toxic.
We use the coarse version of the dataset, following \citet{idrissi2021simple}.
The spurious attribute $s$ determines whether or not the comment mentions one of the following protected attributes: male, female, LGBT, black, white, Christian, Muslim, other religions.
These protected attributes are mentioned more frequently in toxic comments compared to neutral comments, constituting a spurious correlation.
There are $4$ groups corresponding to the pairs $(y, s)$.

\textbf{MultiNLI}\quad
The MultiNLI dataset is described in Figure \ref{fig:text_data_desc}. 
It contains pairs of sentences, and the label $y$ describes the relationship between the sentences:
contradiction, entailment or neutral.
The spurious attribute $s$ describes the presence of negation words, which appear much more frequently in the examples from the negation class.
There are $6$ groups corresponding to the pairs $(y, s)$.

\subsection{The nature of the spurious correlations}
\label{sec:data_discussion}

The nature of the spurious correlation differs between the datasets that we consider.
On Waterbirds, CelebA, Civil Comments and MultiNLI, the spurious attribute is correlated with the target on train, but is not generally predictive of the class label across the groups.
In particular, we wish to train a model that ignores the spurious attribute in its predictions, as using the spurious attribute \textit{hurts} the performance on some of the groups.
By applying DFR on a group-balanced validation set, we try to train a model that ignores the spurious feature.

On the FMOW dataset, the groups correspond to the spurious attribute (region), and not pairs $(y, s)$.
As the regions are not represented equally, standard ERM is incentivised to perform well on the majority groups, with less weight on the minority groups.
In this case, we do not wish to remove the reliance on the spurious attribute, but instead we wish to find a model that performs well on the minority groups.
We apply DFR to this dataset by retraining the last layer on a group-balanced validation set, analogously to the other datasets.

Finally, in the CXR dataset, there are no examples showing healthy patients with a chest drain (to the best of our knowledge).
Consequently, the reliance on the spurious attribute $s$ does not necessarily \textit{hurt} performance on the other groups, unlike e.g. on the Waterbirds dataset.
In this case, we do not wish to remove ignore the feature $s$, but instead we wish to find a model that performs well on the images with $s = 0$ (no chest drain).
Applying DFR to this dataset is not straightforward, as (1) we do not have examples from the $(y=0, s=1)$ group, and (2) we do not wish to remove the reliance on the $s$ attribute.
We found that the best approach in this case was to simply train a logistic regression model on all of the available validation data, without any balancing.
DFR generally provides a smaller improvement on this dataset, and behaves less consistently, as we see e.g. in Figure \ref{fig:architecture}.

\FloatBarrier

\begin{figure}[h]
\begin{lstlisting}[language=Python]
import torchvision.transforms as transforms
import torch


target_resolution = (224, 224)
resize_resolution = (256, 256)
IMAGENET_STATS = ([0.485, 0.456, 0.406], [0.229, 0.224, 0.225])


# No Augmentation
noaug_transform = transforms.Compose(
    transforms.RandomResizedCrop(
        target_resolution,
        scale=(0.7, 1.0),
        ratio=(0.75, 1.33),
        interpolation=2),
    transforms.RandomHorizontalFlip(),
    transforms.ToTensor(),
    transforms.Normalize(*IMAGENET_STATS)
    )

# Default Augmentation
aug_transform = transforms.Compose(
    transforms.Resize(resize_resolution),
    transforms.CenterCrop(target_resolution)
    transforms.ToTensor(),
    transforms.Normalize(*IMAGENET_STATS)
    )

# On test, we always use noaug_transform
test_transform = noaug_transform

\end{lstlisting}
\caption{\textbf{Data augmentation policies.}
The default data augmentation policies implemented using the \lstinline{torchvision} package.}
\label{fig:augmentations_code}
\end{figure}

\subsection{Data augmentation and preprocessing}
\label{sec:app_aug}

On the image datasets, we consider several data augmentation policies.
In Figure \ref{fig:augmentations_code}, we provide the code implementing the \textbf{No augmentation} and \textbf{Default augmentation} policies in torchvision.
The other policies are more involved, so we do not include the code here.
All policies normalize the data analogously to the code in Figure \ref{fig:augmentations_code}.

\textbf{No augmentation.}\quad
This policy resizes the data to a fixed resolution and applies channel normalization.

\textbf{Default policy.}\quad
This policy additionally applies random crops and horizontal flips.

\textbf{MixUp.}\quad
For this policy, we use the Default policy to initially preprocess the images, and then apply MixUp \citep{zhang2017mixup} with the mixing parameter $\alpha = 0.2$.

\textbf{Random Erasing.}\quad
This policy is described in \citet{zhong2020random}.
It randomly erases rectangular blocks of the image, and replaces them with uniform grey blocks.
We use the implementation in \lstinline{timm.data.random_erasing} in the \lstinline{timm} package.

\textbf{Augmix.}\quad
We adapt the official implementation of the AugMix policy \citep{hendrycks2019augmix} available \uhref{https://github.com/google-research/augmix/blob/master/augmentations.py}{here}.

\textbf{Text models.}\quad
For data preprocessing in the experiments on text data, we use the BERT tokenizer: \lstinline{BertTokenizer.from_pretrained("bert-base-uncased")} from the \lstinline{transformers} package.

\subsection{Models}
\label{sec:app_models}

Here, we list the models used in this paper.

\textbf{ResNet-50.}\quad
On the Waterbirds, CelebA and FMOW by default we use the ResNet-50 model \citet{he2016deep} pretrained on ImageNet.
We use the model implemented in the \lstinline{torchvision} package: \lstinline{torchvision.models.resnet50(pretrained=True)}.
In Figure \ref{fig:pretraining_aug}, we additionally consider the randomly initialized model \lstinline{torchvision.models.resnet50(pretrained=False)},
and the models pre-trained with SimCLR \citep{chen2020simple} and Barlow Twins \citep{zbontar2021barlow} contrastive learning methods, imported from the
\lstinline{vissl} package (models available \uhref{https://github.com/facebookresearch/vissl/blob/main/MODEL_ZOO.md}{here}).

\textbf{DenseNet-121.}\quad
On CXR, by default we use the DenseNet-121 model \citep{huang2017densely} implemented in the \lstinline{torchvision} package: \lstinline{torchvision.models.densenet121(pretrained=True)}.
We also consider this model without ImageNet pretraining: \lstinline{torchvision.models.densenet121(pretrained=False)}.

\textbf{Other image models.}\quad
In Section \ref{sec:architectures}, we additionally consider a broad range of architectures and pretraining methods, which we briefly list here.
We use the following models from \lstinline{torchvision}, all pretrained on ImageNet1k: 
ResNet-18, ResNet-34, ResNet-50, ResNet-101, ResNet-152 \citep{he2016deep};
Wide-ResNet-50-2 \citep{zagoruyko2016wide};
ResNext-50-$32\times4d$ \citep{xie2017aggregated};
DenseNet-121 \citep{huang2017densely};
VGG-16, VGG-19 \citep{simonyan2014very};
AlexNet \citep{krizhevsky2014one}.
We also use the following models from the \lstinline{timm} package:
ConvNext-Small, ConvNext-Base, ConvNext-Large, ConvNext-XLarge \citep{liu2022convnet}, pretrained on either ImageNet1k or ImageNet22k;
ViT-Small, ViT-Base, ViT-Large, ViT-Huge \citep{dosovitskiy2020image}, pretrained on either ImageNet1k or ImageNet22k;
BEiT-Base, BEiT-Large \citep{bao2021beit}, pretrained on either ImageNet1k or ImageNet22k;
DEiT-Small, DEiT-Base \citep{bao2021beit}, pretrained ImageNet1k.
We also use ViT-Small, ViT-Base models with DINO pretraining on ImageNet1k \citep{caron2021emerging} available \uhref{https://github.com/facebookresearch/dino}{here}.
Finally, we use ViT-Base, ViT-Large and ViT-Huge models with MAE pretraining on ImageNet1k \citep{he2021masked}, with or without supervised finetuning on ImageNet1k, available \uhref{https://github.com/facebookresearch/mae}{here}.

\textbf{BERT model.}\quad
On the text classification problems, we use the BERT for classification model from the \lstinline{transformers} package:
\lstinline{BertForSequenceClassification.from_pretrained('bert-base-uncased', num_labels=num_classes)}.

\begin{table}
    \footnotesize
    \hspace{-.5cm}
    \begin{tabular}{c cccccc}
\hline
\\[-2mm]
\textbf{Method} & \textbf{Waterbirds} & \textbf{CelebA} & \textbf{FMOW} & \textbf{CXR} & \textbf{MultiNLI} & \textbf{CivilComments} \\
\hline
\\[-2mm]
\textbf{ ERM } & $68.9_{\pm 2.0 }$ & $44.0_{\pm 2.1 }$ & $31.4_{\pm 0.7 }$ & $68.8_{\pm 0.2 }$ & $67.5_{\pm 1.1 }$ & $61.0_{\pm 0.3 }$ \\
\textbf{ ERM + DFR } & $91.1_{\pm 0.8 }$ & $89.4_{\pm 0.9 }$ & $41.6_{\pm 0.6 }$ & $71.6_{\pm 0.5 }$ & $72.6_{\pm 0.3 }$ & $78.8_{\pm 0.5 }$ \\[2mm]
\textbf{ RWY } & $65.4_{\pm 0.6 }$ & $46.1_{\pm 2.1 }$ & $30.5_{\pm 0.6 }$ & $71.3_{\pm 1.1 }$ & $68.0_{\pm 0.4 }$ & $63.4_{\pm 0.9 }$ \\
\textbf{ RWY + DFR } & $90.4_{\pm 1.0 }$ & $88.3_{\pm 0.5 }$ & $40.9_{\pm 0.7 }$ & $72.5_{\pm 0.1 }$ & $72.2_{\pm 1.9 }$ & $78.5_{\pm 0.4 }$ \\[2mm]
\textbf{ RWY-ES } & $74.5_{\pm 0.0 }$ & $76.8_{\pm 7.7 }$ &  &  &  & $78.9_{\pm 1.0 }$ \\
\textbf{ RWY-ES + DFR } & $89.1_{\pm 0.7 }$ & $89.6_{\pm 0.5 }$ &  &  &  & $76.9_{\pm 0.6 }$ \\[2mm]
\textbf{ RWG } & $67.7_{\pm 0.7 }$ & $49.1_{\pm 0.9 }$ & $31.2_{\pm 0.1 }$ &  & $62.0_{\pm 0.2 }$ & $73.5_{\pm 2.2 }$ \\
\textbf{ RWG + DFR } & $91.3_{\pm 0.3 }$ & $85.4_{\pm 1.5 }$ & $41.1_{\pm 0.6 }$ &  & $71.6_{\pm 1.3 }$ & $79.3_{\pm 0.5 }$ \\[2mm]
\textbf{ RWG-ES } & $77.4_{\pm 0.0 }$ & $83.7_{\pm 0.7 }$ &  &  &  & $74.7_{\pm 6.7 }$ \\
\textbf{ RWG-ES + DFR } & $90.7_{\pm 0.2 }$ & $89.8_{\pm 0.3 }$ &  &  &  & $77.0_{\pm 0.2 }$ \\[2mm]
\textbf{ GDRO } & $68.5_{\pm 6.0 }$ & $66.3_{\pm 7.8 }$ & $30.2$ &  & $70.1$ & $70.6$ \\
\textbf{ GDRO + DFR } & $88.2_{\pm 1.1 }$ & $90.4_{\pm 0.7 }$ & $40.3$ &  & $71.8$ & $80.2$ \\[2mm]
\textbf{ GDRO-ES } & $90.7_{\pm 0.6 }$ & $90.6_{\pm 1.6 }$ & $33.1$ &  & $73.5$ & $80.4$ \\
\textbf{ GDRO-ES + DFR } & $89.9_{\pm 0.5 }$ & $91.1_{\pm 0.1 }$ & $42.5$ &  & $73.3$ & $77.3$ \\[2mm]
\hline
\\[0mm]
    \end{tabular}
    \caption{
    \textbf{Method comparison results.}
    Detailed results for the method comparison presented in Figure \ref{fig:methods}.
    The error bars represent one standard deviation over $3$ independent runs.
    We only evaluate early stopping (ES) on datasets where it is helpful for the base model performance.
    On CXR, no group information is available on the train data, so we can only apply the ERM and RWY methods.
    We use a ResNet50 model pretrained on ImageNet on Waterbirds, CelebA and FMOW, a DenseNet-121 model pretrained on ImageNet on 
    CXR, and a BERT model pretrained on Book Corpus and English Wikipedia data on CivilComments and MultiNLI.
    }
    \label{tab:method_results}
\end{table}

\section{Details: ERM vs Group Robustness}
\label{sec:app_methods}

\subsection{Methods and hyper-parameters}

\textbf{ERM, RWY and RWG.} \quad
We report the hyper-parameters used on each of the datasets in Table \ref{tab:erm_hyper}.
We did not tune the hyper-parameters for ERM, RWY and RWG aside from the learning rate for the text classification problems.
For all vision datasets, we used the default data augmentation policy (see Section \ref{sec:app_aug}).
We describe the default model choices in Section \ref{sec:app_models}.
The RWY method is implemented by providing a \lstinline{sampler} to the \lstinline{DataLoader} in \lstinline{PyTorch}, which samples the datapoints from different classes with the same frequency.
The RWG method is implemented analogously, and samples datapoints from different groups with the same frequency.

\begin{table}
\footnotesize
    \hspace{-1cm}
    \begin{tabular}{c c c c c c c}
    \hline \\[-2mm]
    \textbf{Dataset} & \textbf{Optimizer} & \textbf{Initial LR} & \textbf{LR schedule} & \textbf{Batch size} & \textbf{Weight decay} & \textbf{\# Epochs} \\
    \hline \\[-2mm]
    \textbf{Waterbirds} & SGD \citep{robbins1951stochastic}  & $3 \cdot 10^{-3}$ & Cosine annealing & 32 & $10^{-4}$ & 100\\
    \textbf{CelebA} & SGD \citep{robbins1951stochastic}  & $3 \cdot 10^{-3}$ & Cosine annealing & 100 & $10^{-4}$ & 20\\
    \textbf{FMOW} & SGD \citep{robbins1951stochastic}  & $3 \cdot 10^{-3}$ & Cosine annealing & 100 & $10^{-4}$ & 20\\
    \textbf{CXR} & SGD \citep{robbins1951stochastic}  & $3 \cdot 10^{-3}$ & Cosine annealing & 100 & $10^{-4}$ & 20\\
    \textbf{MultiNLI} & AdamW \citep{loshchilov2017decoupled}  & $10^{-5}$ & Linear annealing & 16 & $10^{-4}$ & 10\\
    \textbf{Civil Comments} & AdamW \citep{loshchilov2017decoupled}  & $10^{-5}$ & Linear annealing & 16 & $10^{-4}$ & 10\\
    \hline
    \\[0mm]
    \end{tabular}
    \caption{\textbf{ERM, RWY and RWG hyper-parameters.}
    Default hyper-parameters used on each dataset.
    On the image classification datasets, we adapted the hyper-parameters of \citet{kirichenko2022last}, with no tuning.
    On the text classification dataset, we followed \citet{sagawa2019distributionally} and \citet{idrissi2021simple} in the choice
    of the optimizer and learning rate scheduler, and chose the learning rate value which provided the best base model validation mean accuracy.
    }
    \label{tab:erm_hyper}
\end{table}

\textbf{Group DRO.} \quad
The training objective used in \citet{sagawa2019distributionally} is the following:

$$ \hat{\theta} = \arg \min_{\theta} \max_{g \in \mathcal{G}} \left[ \mathbb{E}_{(x, y) \sim p_g} l(y, f_{\theta}(x)) + C / \sqrt{n_g} \right], $$

where $f_{\theta}(\cdot)$ is a neural network model with parameters $\theta$, $l(\cdot, \cdot)$ is a loss function (cross-entropy for classification), $\mathcal{G}$ is a set of all groups, $n_g$ is the size of the group $g$, and $C$ is a generalization adjustment hyper-parameter.
We follow \citet{sagawa2019distributionally} to choose hyper-parameter combinations for tuning group DRO on Waterbirds, CelebA and MultiNLI.

In particular, on Waterbirds we considered the following combinations of the initial learning rate $lr$ and weight decay $wd$:
${(lr = 10^{-3}, wd=10^{-4}), (lr=10^{-4}, wd=0.1)}$ and $(lr=10^{-5}, wd=1)$. 
We varied the parameter $C$ in the range $\{0, 1, 2, 3, 4, 5\}$ for each combination of learning rate and weight decay. Additionally, we varied weight decay in range
$\{0, 10^{-4}, 10^{-4}, {3 \cdot 10^{-4}}, 10^{-3}, {3 \cdot 10^{-3}}, 10^{-2}, {3 \cdot 10^{-2}}, 0.1, 0.3, 1 \}$ for fixed values of learning rate $10^{-3}$ and $C=0$.

On CelebA, we considered the following combinations of $lr$ and $wd$:
${(lr=10^{-4}, wd=10^{-4})}, {(lr=10^{-4}, wd=10^{-2})}$ and ${(lr=10^{-5}, wd=0.1)}$, varying $C$ in the same range as on Waterbirds. For weight decay ablation we considered values in $
\{0, 10^{-5}, 3 \cdot 10^{-5}, 10^{-4}, 3 \cdot 10^{-4}, 10^{-3}, 3 \cdot 10^{-3}, 10^{-2}\}$.

On MultiNLI, we use learning rate $2 \cdot 10^{-5}$, and vary weight decay in the range $\{0., 0.01, 0.1, 1.0\}$ and $C$ in the range $\{0, 1, 3, 5\}$.

On CivilComments, we considered the following combinations of $lr$ and $wd$:
${(lr=10^{-5}, wd=0.1)}, {(lr=10^{-5}, wd=0.01)}, {(lr=10^{-4}, wd=0.01)}$; for each combination we varied $C$ in $\{0, 3, 5\}$.

On FMOW, the learning rate and weight decay pairs were
${(lr = 10^{-3}, wd = 10^{-3})}$, ${(lr = 10^{-4},  wd = 10^{-3})}$, ${(lr = 10^{-4},  wd = 10^{-2})}$, ${(lr = 10^{-5},  wd = 10^{-1})}$. For each $lr$ and $wd$ combination we varied $C$ in the same range as on Waterbirds and CelebA datasets.

For all image datasets, we use the default data augmentation policy (see Section \ref{sec:app_aug}). In all runs we used the same batch size and train for the same number of epochs as the corresponding hyper-parameters in ERM (see Table \ref{tab:erm_hyper}).

\subsection{DFR implementation and hyper-parameters}

In all experiments, we use the \DFRVAL variation of the DFR method described in \citet{kirichenko2022last}.
We follow the official implementation provided \uhref{https://github.com/PolinaKirichenko/deep_feature_reweighting}{here}.
Specifically, we use $\ell_1$ regularization for training the logistic regression model implemented in the \lstinline{scikit-learn} package:
\lstinline{sklearn.LogisticRegression(penalty="l1", C=c, solver="liblinear")};
we tune the regularization strength $c$ within the set $\{1, 0.7, 0.3, 0.1, 0.07, 0.03, 0.01\}$, following the procedure described in \citet{kirichenko2022last}.

As explained in Section \ref{sec:data_discussion}, on CXR we train the logistic regression model on all of the validation set without group balancing.
Further, on CXR we tune the regularization strength parameter according to the worst AUC, and not worst group accuracy.

We additionally compute \DFRS\ by using DFR (with the same hyper-parameters and implementation) to predict the spurious attribute $s$ (instead of the class label $y$) from the learned features.

\subsection{Results}

We provide detailed results for all methods on all datasets in Table \ref{tab:method_results}.
Group-DRO  significantly improves the base model performance compared to all other methods across the board.
After applying DFR, the performance across the different methods is very similar, although Group-DRO with early stopping still typically provides a small improvement.
The improvement is, however, very small compared to the improvement from using a better base model (see Section \ref{sec:architectures}).

\section{Details: Effect of the Base Model}
\label{sec:app_arch_ablation}

For all experiments in Section \ref{sec:architectures}, we use the default data augmentation policy (see Section \ref{sec:app_aug}).
We consider a broad range of models and architectures (see Section \ref{sec:app_models}).
We ran all models with the default hyper-parameters provided in Table \ref{tab:erm_hyper}.
For CXR-14 dataset, we used class reweighting due to heavy class imbalance present in the train data ($95\%$ of train images are from the negative class).

We note that the default hyper-parameters are suboptimal for some of the ViT models, which lead to poor performance for some of the models.
With more tuning, we expect that it should be possible to improve the results further for all the considered models, especially ViT-based.

\section{Details: Effect of Regularization}
\label{sec:app_regularization}

\textbf{Effect of weight decay.}\quad
For the experiments on the effect of weight decay we use the default models described in Section \ref{sec:app_models}, and default hyper-parameters in Table \ref{tab:erm_hyper},
and vary the weight decay strength in the range $\{0, 10^{-5}, 10^{-4}, 3\cdot 10^{-4}, 10^{-3}, 3 \cdot 10^{-3}, 10^{-2}\}$.
We report the results in Figure \ref{fig:wd_effect}.

\textbf{Effect of data augmentation.}\quad
For the experiments on the effect of data augmentation, we use the default models described in Section \ref{sec:app_models}, and default hyper-parameters in Table \ref{tab:erm_hyper},
and apply the data augmentation policies desribed in \ref{sec:app_aug}.
We report the results in Figure \ref{fig:pretraining_aug}(b).

\begin{figure}[t]
\centering
\includegraphics[width=0.99\textwidth]{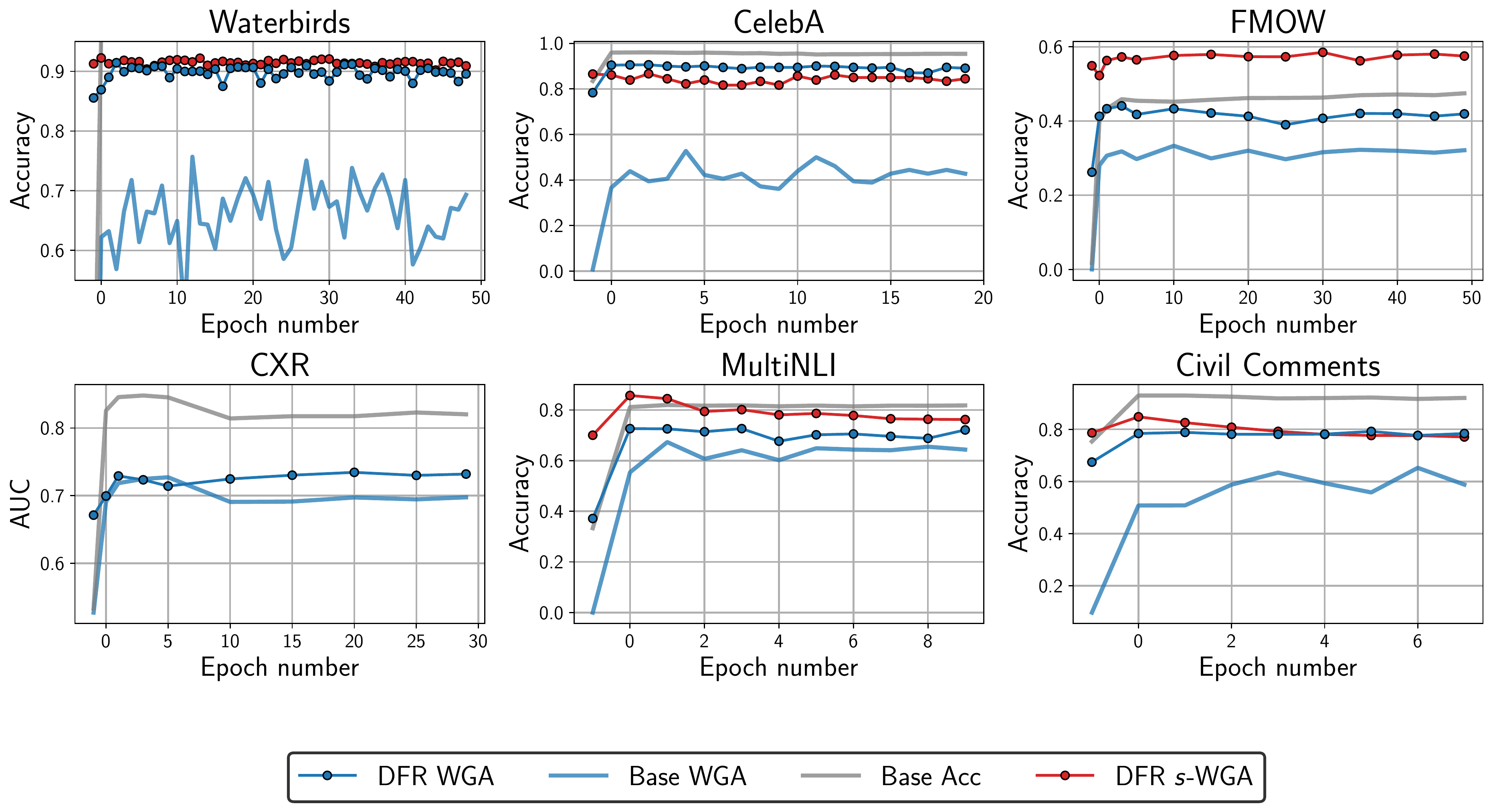}
\caption{
\textbf{ERM training length effect.}
On all datasets, $5$ epochs or less is sufficient to achieve near-optimal DFR WGA performance,
but longer training does not hurt performance.
}
\label{fig:app_trainlen}
\end{figure}

\textbf{Effect of training length on ERM.}\quad
We plot the DFR WGA, \DFRS, as well as base model WGA and mean accuracy as a function of training epoch for ERM training in Figure \ref{fig:app_trainlen}.
On all datasets, we observe that the DFR WGA quickly converges and stays roughly constant throughout training.
On all datasets, $5$ epochs is sufficient for near-optimal performance.
Longer training generally does not help or hurt DFR WGA, even when it hurts the base model.

\begin{figure}[t]
\centering
\includegraphics[width=0.99\textwidth]{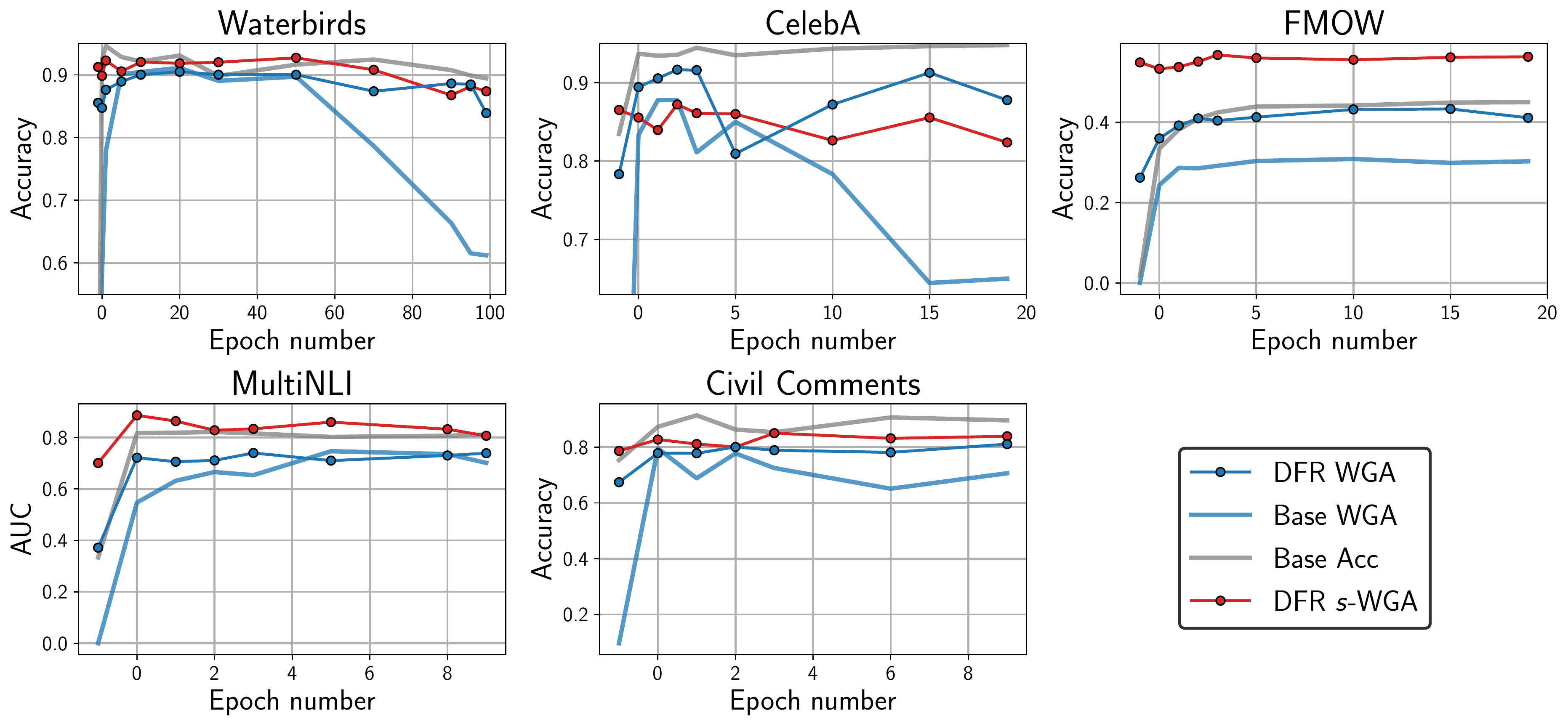}
\caption{
\textbf{Group-DRO training length effect.}
DFR WGA converges to near-optimal performance in under $5$ epochs.
On Waterbirds and CelebA the DFR WGA performance is less stable compared to the ERM results in Figure \ref{fig:app_trainlen}.
}
\label{fig:app_trainlen_gdro}
\end{figure}

\textbf{Effect of training length on Group-DRO.}\quad
We repeat the same experiment, but with Group-DRO instead of ERM training in Figure \ref{fig:app_trainlen_gdro}.
Again, we observe that $5$ epochs are generally sufficient for near-optimal performance. 
Interestingly, for Group-DRO, DFR WGA does deteriorate over time on Waterbirds but not as significantly as the base model WGA.

\begin{table}[t]
\small
    \centering
\begin{tabular}{@{}lllll@{}}
\toprule
\textbf{Base Model} & \textbf{Base Acc} & \textbf{Base WGA} & \textbf{DFR WGA} & \textbf{DFR s-WGA}\\ 
\midrule
\textbf{BERT-Base} & {82.3} & {66.1} & {73.8} & {83.6} \\
\textbf{BERT-Large} & 84.6 & 70.6 & 76.6 & 79.0 \\
\textbf{DeBERTa-Base} & 88.9 & 80.4 & 82.3 & 80.2 \\
\textbf{DeBERTa-Large} & 90.2 & 81.2 & 84.8 & 68.3 \\
\bottomrule\\
\end{tabular}
\caption{\textbf{MultiNLI: base model effect.}
Effect of base model on the performance on the MultiNLI dataset.
DeBERTa-Large model provides the best performance in terms of DFR WGA as well as Base WGA and Base mean accuracy.}
\label{tab:multinli_arch}
\end{table}

\begin{table}[t]
\footnotesize
    \centering
\begin{tabular}{@{}cllll@{}}
\toprule
\textbf{Pretraining} & \textbf{Base Acc} & \textbf{Base WGA} & \textbf{DFR WGA} & \textbf{DFR s-WGA}\\ 
\midrule
\textbf{Random Init} & 59.1 & 26.0 & 45.4 & 94.9 \\
\textbf{Pretrained on Wiki + Book Corpus} & 82.3 & 66.1 & 73.8 & 83.6 \\
\textbf{MultiLingual on Wiki} & 80.5 & 62.1 & 72 & 83.9 \\
\bottomrule\\
\end{tabular}
\caption{\textbf{MultiNLI: effect of pretraining.}
Results for BERT-Base model with different types of pre-training on the MultiNLI dataset.
Pretraining is required to achieve strong performance.
Multilingual pretraining is competitive but inferior to pretraining on data in English.
}
\label{tab:multinli_pretraining}
\end{table}

\begin{table}[t]
\footnotesize
    \hspace{-0.7cm}
\begin{tabular}{@{}cllll@{}}
\toprule
\textbf{Pretraining} & \textbf{Init DFR WGA} & \textbf{Trained DFR WGA} & \textbf{Init DFR s-WGA} & \textbf{Trained DFR s-WGA}\\ 
\midrule
\textbf{BERT-Base} & 37.4 & 73.8 & 69.9 & 83.6 \\
\textbf{BERT-Large} & 34.4 & 76.6 & 57.6 & 81 \\
\textbf{DeBERTa-Base} & 58.5 & 82.3 & 77 & 80 \\
\textbf{DeBERTa-Large} & 61.8 & 84.8 & 66.4 & 68.3 \\
\bottomrule\\
\end{tabular}
\caption{\textbf{MultiNLI: effect of training on target data.}
Results for different models before and after training on the MultiNLI dataset.
For all the considered models, both the core and the spurious features are significantly more decodabe after training on the target data.
}
\label{tab:multinli_target}
\end{table}

\begin{table}[t]
\small
    \centering
\begin{tabular}{@{}cllll@{}}
\toprule
\textbf{Weight decay} & \textbf{Base Acc} & \textbf{Base WGA} & \textbf{DFR WGA} & \textbf{DFR s-WGA}\\ 
\midrule
\textbf{0} & 82.3 & 66.1 & 73.8 & 83.6\\
\textbf{1} & 81.4 & 63.2 & 74.6 & 90.1\\
\textbf{3} & 74.5 & 43.0 & 66.0 & 94.4\\
\textbf{10} & 63.5 & 23.4 & 39.9 & 66.0\\
\textbf{30} & 57.0 & 15.2 & 28.9 & 59.3\\
\textbf{100} & 46.4 & 1.7 & 3.2 & 53.0 \\
\bottomrule\\
\end{tabular}
\caption{\textbf{MultiNLI: weight decay.}
The effect of weight decay on the BERT-Base model on MultiNLI.
Similarly to the results in Figure \ref{fig:wd_effect}, weight decay $0$ provides
competitive performance.
The best DFR WGA is achieved with weight decay $1$.
}
\label{tab:multinli_wd}
\end{table}

\section{Additional results on MultiNLI}
\label{sec:app_multnli_res}

For all experiments in this section, we train the models for $5$ epochs with learning rate $10^{-5}$ and $0$ weight decay.

\textbf{Effect of base model.}\quad In Table \ref{tab:multinli_arch}, we report the results for \uhref{https://huggingface.co/bert-base-uncased}{BERT-Base}, \uhref{https://huggingface.co/bert-large-uncased}{BERT-Large}, 
\uhref{https://huggingface.co/microsoft/mdeberta-v3-base}{DeBERTa-Base},
and
\uhref{https://huggingface.co/microsoft/deberta-v3-large}{DeBERTa-Large} models \citep{he2021debertav3, he2020deberta}.
For BERT models, following \citet{sagawa2019distributionally}, we use the cached tokenizer outputs with maximum sequence length $150$.
For DeBERTa models, we re-tokenize the dataset with the corresponding tokenizer with a maximum sequence length of $220$. We train all models for $5$ epochs.
The advanced DeBERTa-Large model provides the best base performance and DFR WGA.

\textbf{Effect of pretraining.}\quad
In Table \ref{tab:multinli_pretraining} we evaluate the results of BERT-Base models with different types of pretraining on MultiNLI.
This experiment is analogous to the experiment for image classification problems presented in Figure \ref{fig:pretraining_aug}(a).
We find that pretraining is necessary to achieve strong performance on this dataset, but different pretraining datasets lead to competitive results.

\textbf{Effect of training on target data.}\quad
In Table \ref{tab:multinli_target} we evaluate the effect of training on the MultiNLI dataset for the BERT and DeBERTa pretrained models.
This experiment is analogous to the experiment for image classification problems presented in Figure \ref{fig:architecture_training}.
We find that after training on the target data both the core features (DFR WGA) and the spurious features (DFR $s$-WGA) become significantly more decodable.
This result is in contrast to the results on Waterbirds in Figure \ref{fig:architecture_training}, where DFR WGA is not significantly improved from training.

\textbf{Effect of weight decay.}\quad 
In Table \ref{tab:multinli_wd} we report the results of the weight decay ablation for the BERT-Base model on MultiNLI; this model uses the AdamW optimizer \citep{loshchilov2017decoupled}, so we consider larger values of weight decay.

\section{Broader Impact and Limitations}
\label{sec:limitations}

\textbf{Limitations.}\quad
While we consider a wide range of factors that affect the feature learning under spurious correlations, we inevitably do not cover all the possible factors.
In particular, it would be interesting to consider the effect of regularization methods beyond weight decay and early stopping, and methods for diverse feature learning such as DivDis \citep{lee2022diversify}, or the method of \citet{teney2021evading}.
As another limitation, while DFR performs well in our experiments, it is not guaranteed to learn an optimal linear classifier with the given features;
further improvements in learning the last layer can be used to refine the results of our study.
Despite these limitations, we believe that our work provides a comprehensive analysis of the feature learning under spurious correlations.

\textbf{Broader impact.}\quad
Research on spurious correlations is closely related to ML Fairness \citep{dwork2012fairness,hardt2016equality, kleinberg2016inherent, pleiss2017fairness, agarwal2018reductions, khani2019maximum}.
We hope that our work can motivate further research in fairness, where techniques similar to DFR can be considered to improve the fairness of ML models.
A potential negative outcome that can result from \textit{misinterpretation} of our analysis is if the practitioners assume that spurious correlations are not an important issue, as ERM learns high quality representation of the core features.
We emphasize that ERM still performs suboptimally (see Figure \ref{fig:methods}), as it does not provide a correct weighting for the features in the final classification layer.
Spurious correlations are a significant practical issue that should be considered carefully in real-world applications.

\textbf{Compute.} \quad
We estimate the total compute used in the process of working on this paper at roughly $3000$ GPU hours.
The compute usage is dominated by the experiments presented in Figure \ref{fig:architecture}, where we trained a large number of large-scale models on $4$ vision datasets.
The tuning of Group-DRO hyper-parameters was also relatively compute-heavy.
The experiments were run on GPU clusters on Nvidia Tesla V100, Titan RTX, RTX8000, 3080 and 1080Ti GPUs.

\textbf{Licenses.} \quad
The Civil Comments dataset is distributed under the \uhref{https://creativecommons.org/publicdomain/zero/1.0/}{CC0} license.
The FMOW dataset is under the \uhref{https://github.com/fMoW/dataset/blob/master/LICENSE}{FMoW Challenge Public License}.
The Places dataset is under the \uhref{https://creativecommons.org/licenses/by/4.0/}{CC BY} license.
For the details of the license for the MultiNLI dataset, see \citet{williams2017broad}.

\end{document}